\documentclass[runningheads]{llncs}

\usepackage{eccv}

\usepackage{eccvabbrv}

\usepackage{graphicx}
\usepackage{booktabs}

\usepackage[accsupp]{axessibility}  

\usepackage{hyperref}

\usepackage{orcidlink}

\usepackage{amsmath}
\usepackage{amssymb}

\usepackage{epsfig} 
\usepackage{color}
\usepackage{multirow}
\usepackage{physics}
\usepackage{tabu}
\usepackage{enumitem}

\newcommand{\PreserveBackslash}[1]{\let\temp=\\#1\let\\=\temp}
\newcolumntype{C}[1]{>{\PreserveBackslash\centering}p{#1}}

\newcommand{\tabincell}[2]{\begin{tabular}{@{}#1@{}}#2\end{tabular}}

\newcommand{\supp}[1]{{\color{black}#1}}

\begin{document}

\title{Generative Hierarchical Temporal Transformer for Hand Pose and Action Modeling} 

\titlerunning{G-HTT}


\author{Yilin Wen\inst{1,2} \orcidlink{0000-0002-5981-1276}
\and Hao Pan\inst{3} \orcidlink{0000-0003-3628-9777} 
\and Takehiko Ohkawa\inst{2}\orcidlink{0000-0003-2329-8797}
\and Lei Yang\inst{1,4}\orcidlink{0000-0002-3284-4019} 
\and Jia Pan\inst{1,4}\orcidlink{0000-0001-9003-2054}
\and Yoichi Sato\inst{2}\orcidlink{0000-0003-0097-4537}
\and Taku Komura\inst{1}\orcidlink{0000-0002-2729-5860} 
\and Wenping Wang\inst{5}\orcidlink{0000-0002-2284-3952}
}
\authorrunning{Y. Wen et al.}
%
\institute{The University of Hong Kong \and The University of Tokyo \and Microsoft Research Asia \and Centre for Garment Production Limited, Hong Kong \and Texas A\&M University}

\maketitle

\begin{abstract}

We present a novel unified framework that concurrently tackles recognition and future prediction for human hand pose and action modeling.
Previous works generally provide isolated solutions for either recognition or prediction, which not only increases the complexity of integration in practical applications, but more importantly, cannot exploit the synergy of both sides and suffer suboptimal performances in their respective domains.
To address this problem, we propose a generative Transformer VAE architecture to model hand pose and action, where the encoder and decoder capture recognition and prediction respectively, and their connection through the VAE bottleneck mandates the learning of consistent hand motion from the past to the future and vice versa.
Furthermore, to faithfully model the semantic dependency and different temporal granularity of hand pose and action, we decompose the framework into two cascaded VAE blocks: the first and latter blocks respectively model the short-span poses and long-span action, and are connected by a mid-level feature representing a sub-second series of hand poses. 
This decomposition into block cascades facilitates capturing both short-term and long-term temporal regularity in pose and action modeling, and enables training two blocks separately to fully utilize datasets with annotations of different temporal granularities. 
We train and evaluate our framework across multiple datasets; results show that our joint modeling of recognition and prediction improves over isolated solutions, and that our semantic and temporal hierarchy facilitates long-term pose and action modeling.


\end{abstract}

\section{Introduction}
Understanding dynamic hand poses and actions is fundamental in fields such as human-robot interaction and VR/AR applications. 
In recent years, huge progress has been made in recognizing 3D hand poses and actions (\eg~take out a chip) from inputs such as RGB videos~\cite{zimmermann2017learning,iqbal2018hand,feichtenhofer2019slowfast,shi2019two,wen2023hierarchical,yang2020collaborative}. 
Meanwhile, another line of research~\cite{bao2023uncertainty,liu2020forecasting,liu2022joint} focuses on predicting future hand motion represented as a sequence of frame-wise poses, where recent research~\cite{lucas2022posegpt} further employs generative models to achieve diverse motion prediction conditioned on a given action. 

\begin{figure}[!t]
\centering
\includegraphics[width=.58\linewidth]{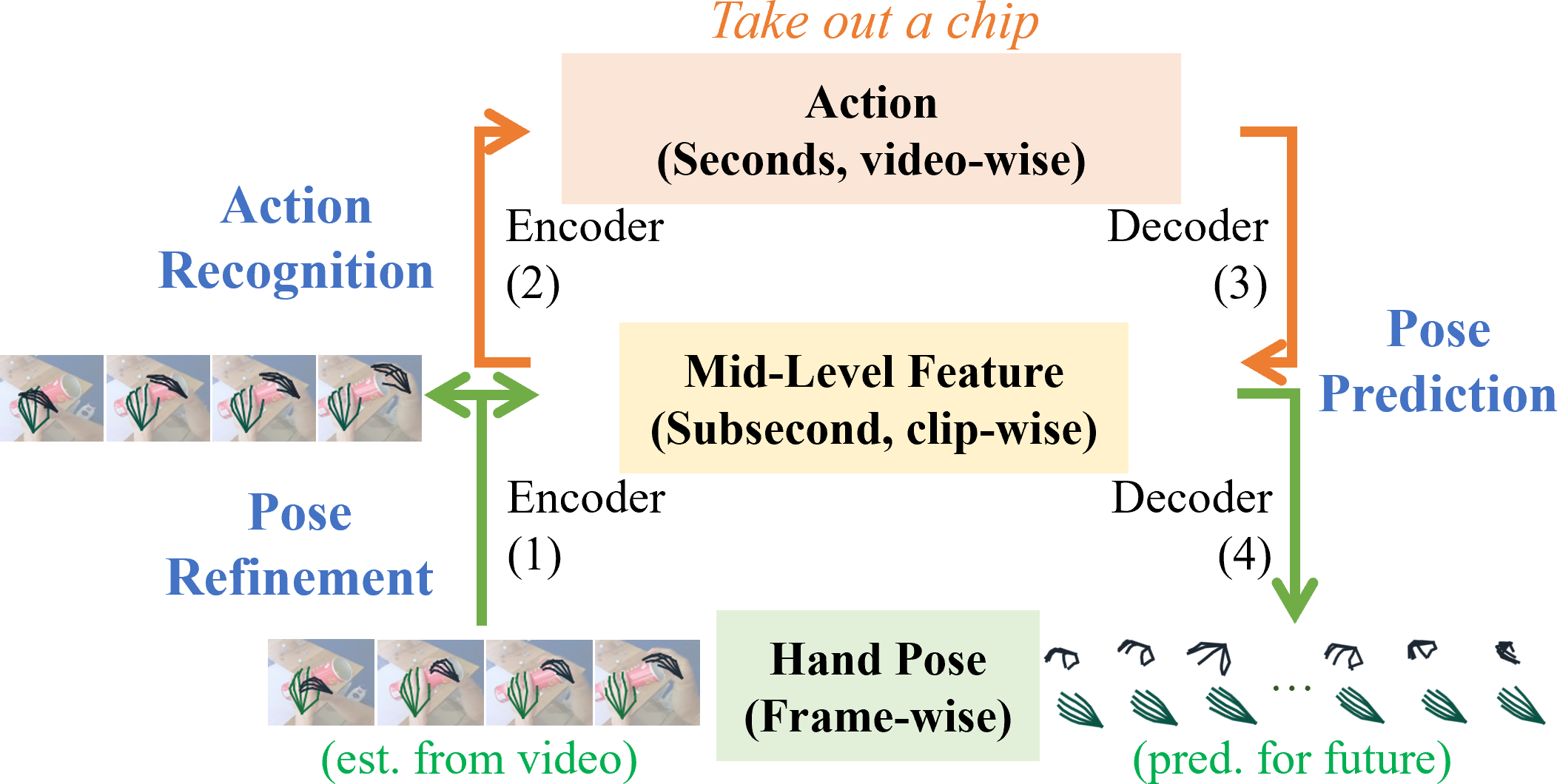}
\caption{Jointly modeling recognition and prediction, while following the semantic dependency and temporal granularity for hand pose-action.
For recognition, (1)$\to$(2) moves up from short to long spans for input pose refinement and action recognition respectively. 
For motion prediction, two paths are available: (1)$\to$(4) exploits short-term motion regularity, and (1)$\to$(2)$\to$(3)$\to$(4) enables long-term action-guided prediction.}\label{fig:hierarchy}
\end{figure}

However, existing literature provides isolated solutions working on either the recognition~\cite{zimmermann2017learning,iqbal2018hand,feichtenhofer2019slowfast,shi2019two,wen2023hierarchical,yang2020collaborative} or prediction~\cite{bao2023uncertainty,liu2020forecasting,liu2022joint,lucas2022posegpt} side, which brings deployment complexity when integrating both sides in practical applications.
Furthermore, we note that recognition and prediction tasks are naturally synergized by the temporal regularities shared through observed and future timestamps: As exemplified in~\cref{fig:hierarchy}, given the observation of hand poses reaching into the can, which indicates the action of taking out a chip, one can predict that future hand motion will describe grabbing and pulling out a chip, thus completing the action. 
However, isolated solutions cannot fully exploit this temporal regularity, resulting in a tendency to overfit to specific data distributions and suffer from suboptimal performances in their respective domains (\cref{sec:exp_jointrp}).

We present a framework with a generative Transformer VAE architecture to jointly capture both recognition and future prediction for hand pose and action modeling, therefore addressing various tasks including input 3D hand pose refinement, action recognition, and future 3D hand motion prediction. 
Our transformer encoder and decoder respectively produce outputs for recognition and prediction, while the VAE latent code connecting the two forces the extraction of regular and consistent hand motion and action, by predicting the future from the past and vice versa. In this way, we synergize recognition and prediction tasks and enhance the performance compared to isolated solutions (\cref{sec:exp_jointrp}).

Moreover, when it comes to pose-action modeling, extensive literature has shown the benefits of capturing the semantic dependency between the instantaneous pose and action over seconds. For example,~\cite{guo2020action2motion, petrovich2021action, petrovich22temos, tevet2022motionclip, lucas2022posegpt} generate motion by conditioning on action to enhance realism. On the recognition side,~\cite{wen2023hierarchical,cho2023transformer,yang2020collaborative,tekin2019h+} aggregates the frame-wise hand poses estimated throughout the video to recognize the performed action. Besides modeling the semantic dependency, Wen~\etal~\cite{wen2023hierarchical} further emphasize capturing the different temporal granularity of hand pose and action. Their framework, namely \textit{Hierarchical Temporal Transformer}, has two cascaded encoders that capture short and long time spans respectively for effective hand pose estimation and action recognition.

In addition to our joint modeling of both recognition and prediction, we are inspired by~\cite{wen2023hierarchical} to introduce block cascades, thus faithfully respecting the semantic dependency and temporal granularity of hand pose-action.
To this end, we decompose our framework into two cascaded blocks that have the same generative transformer VAE structure but focus on different semantic and temporal granularities, thus naming our framework \textit{Generative Hierarchical Temporal Transformer} (G-HTT): The lower pose block (\textbf{P} block) models hand poses over short time spans, and the upper action block (\textbf{A} block) models action over long time spans. A middle-level representation is further introduced to connect the two blocks: it is simultaneously the pose block VAE latent code and the action block encoder input/decoder output, and semantically encodes clip-wise motion over a subsecond span (\cref{fig:hierarchy}).

This decomposition into block cascades offers two key advantages: First, it decouples the complex motion generation into hierarchical subtasks to respectively capture short-term and long-term temporal regularity, improving over flattened models (\cref{sec:exp_hierarchy}). Second, it brings training flexibility, as we can train the blocks separately, which not only reduces training computational cost but also allows for using datasets of different annotation granularities (\cref{sec:method_training,sec:exp_others}).

We train and evaluate the framework across different datasets of two-hand interactions, including H2O~\cite{kwon2021h2o} for daily activities, Assembly101~\cite{sener2022assembly101} and AssemblyHands~\cite{ohkawa2023assemblyhands} for (dis-)assembling take-part toys. 
At test time, given a 3D hand pose sequence (\eg per-frame estimations from the observed RGB video), we first refine it by leveraging the short-term hand motion regularity, (\cref{fig:hierarchy}, (1)). Next, we aggregate the clip-wise motions for action recognition (\cref{fig:hierarchy}, (1)$\to$(2)). Finally, we decode the observed motions and action into a sequence of future middle-level features for motion prediction (\cref{fig:hierarchy}, (1)$\to$(2)$\to$(3)$\to$(4)). 
Evaluation results across datasets show that our framework can solve recognition problems from various camera views, and generate plausible future hand poses over time. 
The contribution of this paper can be summarized as follows: 
\begin{itemize}
\itemsep0em
\item A generative Transformer VAE architecture to concurrently capture recognition and future prediction for hand pose and action modeling, which exploits the temporal regularity synergized between the past and the future, thus improving over isolated solutions.
\item A hierarchical architecture composed of two cascaded generative blocks, which models semantic dependency and temporal granularity of pose-action. This block cascade facilitates capturing both short-term and long-term temporal regularities, and further brings training flexibility.
\item A comprehensive evaluation of the system on tasks such as 3D hand pose refinement, action recognition, and 3D hand motion prediction, validating the performance and design of our framework.

\end{itemize}

\section{Related Works}
\subsubsection{Action Recognition and 3D Hand Pose Estimation}
Massive literature addresses perceiving hand pose and action from visual observation.
For example, a series of works aims to recover the 3D hand skeleton or mesh from the visual input, where the spatial correlation within a single-frame is well exploited~\cite{zimmermann2017learning,iqbal2018hand,spurr2020weakly,li2022interacting,moon2023bringing,yu2023acr}, and the motion coherence along the temporal dimension is further leveraged to improve robustness under occlusion and truncation~\cite{mueller2018ganerated,cai2019exploiting,fan2020adaptive,han2020megatrack,wang2020rgb2hands,han2022umetrack}. 
Meanwhile, \cite{carreira2017quo,feichtenhofer2016convolutional,feichtenhofer2019slowfast,feichtenhofer2020x3d,shi2019two} focus on the higher semantic level, where they extract the spatial-temporal feature from the input frames to recognize the semantic hand or body action.

Moreover, many works notice and exploit the benefits of modeling the semantic dependency between hand pose and action, since intuitively action is defined by the pattern of hand motion (\ie verb) and object in manipulation (\ie noun). 
For example, \cite{li2015delving,ma2016going,singh2016first,tekin2019h+,yang2020collaborative,wen2023hierarchical,cho2023transformer} leverage the hand pose features for action recognition, 
while Yang~\etal~\cite{yang2020collaborative} further refer to the action feature for pose refinement. 
Wen~\etal~\cite{wen2023hierarchical} further stress capturing the respective temporal granularity of pose and action when exploiting temporal cues, and propose a framework with two cascaded blocks to respectively work on short- and long-term spans and output per-frame 3D hand pose and video action.

The hierarchical structure of our framework is inspired by~\cite{wen2023hierarchical}, but we have extended it to model prediction tasks, which not only covers more tasks but also enhances recognition performance (\cref{sec:exp_jointrp}).

\subsubsection{3D Human Hand and Body Motion Prediction} 
Previous works predict the 2D or 3D trajectory of hand roots~\cite{liu2022joint,liu2020forecasting,luo2019human,bao2023uncertainty} or skeleton~\cite{chik2008using} from the observed hand motion. 
Furthermore, \cite{yuan2020dlow,aliakbarian2020stochastic,ma2022multi,cai2021unified,mao2022weakly,tevet2022human,lucas2022posegpt,jiang2023motiongpt,zhang2023motiongpt} capture the distribution of future body motion with powerful generative deep neural networks.
Motion prediction can also benefit from semantic dependency modeling, as achieved by taking the past motion together with a specified action as condition, based on cVAEs~\cite{cai2021unified,mao2022weakly}, GPT-like models~\cite{lucas2022posegpt,jiang2023motiongpt,zhang2023motiongpt} and diffusion models~\cite{tevet2022human}.
For example, PoseGPT~\cite{lucas2022posegpt} first quantizes short motion clips into latent codes by training a VQ-VAE, and then constructs a GPT-like auto-regressive model for motion generation, which learns on sequences of action and latent motion tokens.

Our work builds a hierarchical structure for motion prediction, where consistencies in both short-term motion and long-term action are explicitly ensured through the cascade of generative Transformer VAEs (\cref{sec:exp_hierarchy}).
In addition, we learn prediction and recognition simultaneously, which improves both tasks by exploiting the shared temporal regularity (\cref{sec:exp_jointrp}). 

\subsubsection{Bridging Recognition and Prediction}
There are previous attempts to bridge recognition and prediction, therefore benefiting recognition or prediction at the pose or action level.
For example, \cite{rempe2021humor,shi2023phasemp} learn next-frame pose prediction with cVAEs to model a latent space depicting pose transitions along the temporal dimension. Their learned latent space then serves as a strong regulation in test-time optimization for body pose estimation. 
\cite{petrovich22temos,petrovich2021action,lucas2022posegpt,tevet2022motionclip,tevet2022human} learn text-guided motion generation models, with the text guidance in the form of prescribed actions. The generated sequences can then be used as training data for recognition tasks. 
On the other hand, \cite{gammulle2019predicting,vondrick2016anticipating,zhao2022real,xu2019temporal,chi2023infogcn++} leverage per-frame prediction for understanding high-level action, benefiting tasks such as action anticipation~\cite{gammulle2019predicting,vondrick2016anticipating} or early action detection~\cite{zhao2022real,xu2019temporal,chi2023infogcn++}.

In comparison, our hierarchical modeling enables capturing both recognition and prediction by modeling the semantic hierarchy between short-term pose and long-term action, which significantly boosts computational efficiency, pose/action estimation accuracy, and long-term generation fidelity. This is not considered by existing works.

\section{Methods}

The core framework, namely \textit{Generative Hierarchical Temporal Transformer} (G-HTT, ~\cref{fig:network}), takes as input the object in manipulation and the observed pose sequence of $T$ frames for two interacting hands, where the hand motion and object feature respectively depict the \textit{verb} and \textit{noun} of the action being performed (\eg take out a chip).
G-HTT then jointly models both recognition and prediction, while following the semantic-temporal hierarchy of pose-action that captures their dependency and different temporal granularities (\cref{sec:method_hierahrcy}).
In the test stage, we apply G-HTT to recognition tasks of input pose refinement and action recognition, and to the generation task of diverse hand motion prediction (\cref{sec:method_tasks}).
Important implementation details are given in~\cref{sec:method_training}. \supp{The details of our hand motion representation and a table of notations are provided in the supplementary for reference.}

\subsection{Joint Modeling of Recognition and Prediction with Semantic-Temporal Hierarchy}\label{sec:method_hierahrcy}

\begin{figure*}[!tb]
\centering
\includegraphics[width=0.96\linewidth]{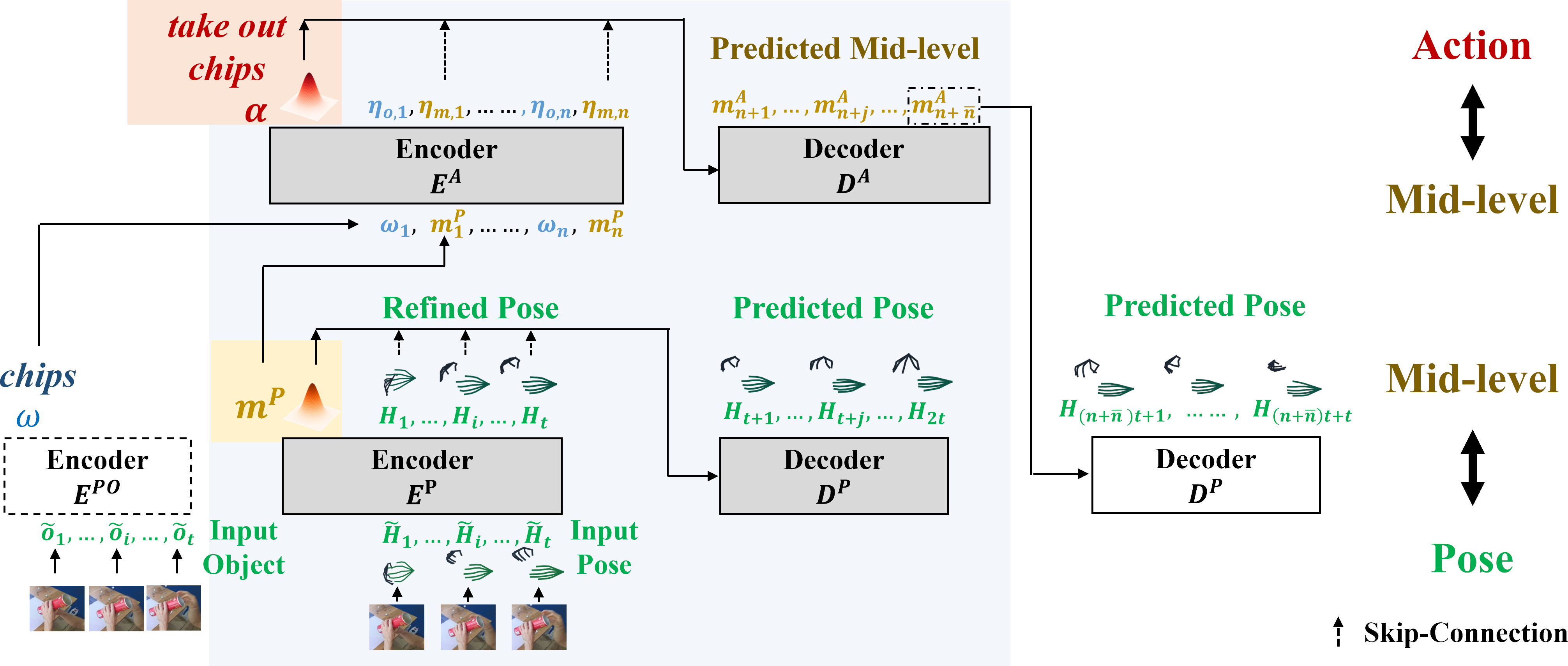}
\caption{Overview of our framework. The cascaded $\vb{P}$ and $\vb{A}$ (shaded in blue) of G-HTT jointly model recognition and prediction, and faithfully respect the semantic dependency and temporal granularity among pose, mid-level and action (\cref{sec:method_hierahrcy}). }\label{fig:network} 
\end{figure*}

G-HTT consists of two cascaded blocks, namely the short-term pose block $\vb{P}$ and the long-term action block $\vb{A}$, to jointly model recognition and prediction while following the hierarchy of temporal and semantic granularity for pose-action.
Both $\vb{P}$ and $\vb{A}$ have the same VAE structure, with their encoders and decoders respectively outputting for recognition and future motion prediction, but $\vb{P}$ and $\vb{A}$ model different semantic levels and time spans (\cref{fig:network}).

To bridge the pose and action blocks in the semantic-temporal hierarchy, we explicitly introduce a mid-level feature $\vb{m}$, which represents the hand poses within a sub-second time span. 
$\vb{P}$ and $\vb{A}$ then respectively model the mappings between pose \vs mid-level, and mid-level \vs action. 
As the two blocks are cascaded, the different semantic levels can refer to each other for globally consistent recognition and prediction (\cref{sec:method_tasks}).
Moreover, our design enables a flexible training scenario, where $\vb{P}$ and $\vb{A}$ can be decoupled and trained separately based on their respective supervision signals and training data (\cref{sec:method_training}).

\subsubsection{P-Block} takes a subsecond time span of $t$ $(t<T)$ consecutive frames to model the relationship between per-frame hand pose and mid-level feature $\vb{m}^P$, without explicitly leveraging the action information.
The mid-level $\vb{m}^P$ is learned to be the latent bottleneck of $\vb{P}$, which encodes the input $t$ consecutive frames of hand poses, and is decoded to hand motion of the future $t$ frames. 
Meanwhile, similar to HTT~\cite{wen2023hierarchical}, the input hand poses can be refined via the encoder $\vb{E}^P$ by leveraging the short-term temporal regularity.

In detail, $\vb{E}^P$ takes as input a sequence of $t+2$ tokens $(\tilde{\mu}^P, \tilde{\Sigma}^P, \tilde{\vb{H}}_1,...,\tilde{\vb{H}}_t)$. $\tilde{\vb{H}}_i$ represents the per-frame hand pose, and $\tilde{\mu}^P, \tilde{\Sigma}^P\in\mathbb{R}^d$ are trainable tokens for parameterizing the distribution of $\vb{m}^P$ by aggregating over $\tilde{\vb{H}}_{1:t}$, similar to~\cite{petrovich22temos,petrovich2021action}. 
Denoting the output sequence of $({\mu}^P, {\Sigma}^P, \vb{H}_1,...,\vb{H}_t)$, we obtain $\vb{H}_i$ as the refined frame-wise hand poses, and sample $\vb{m}^P$ with re-paramterization~\cite{kingma2013auto} from the normal distribution $N({\mu}^P, {\Sigma}^P)$.

$\vb{D}^P$ predicts the hand pose for the following $t$ frames in a parallel manner. 
The mid-level $\vb{m}^P$, with $\vb{H}_{1:t}$ optionally concatenated as skip-connections, are referred by $\vb{D}^P$ through cross-attention. 
As a parallel transformer decoder, the query input of $\vb{D}^P$ are the sinusoidal position encoding of $t$ tokens, and the output $t$ tokens are mapped into the future hand motion $\vb{H}_{t+1:2t}$.

Denoting the corresponding GT motion as $\overline{\vb{H}}_{1:2t}$, $\vb{P}$ is trained by a loss function consisting of three parts: 
\begin{itemize}
    \item The hand component loss to compare the refined and predicted motion with GT:
        \begin{equation}
        L_{comp}=\frac{1}{2t}\sum_{i=1}^{2t}||\overline{\vb{H}}_i-{\vb{H}}_i||_1
        \end{equation}

    \item The frame-wise root trajectory loss for the predicted part:
        \begin{equation}
        L_{trj}=\frac{1}{t}\sum_{i=t+1}^{2t}\left(||\overline{s}^L_i-{s}^L_i||_1+||\overline{s}^R_i-{s}^R_i||_1\right)
        \end{equation}        
    with $\overline{s}^L_i,\overline{s}^R_i$ denote the GT counterpart.
    
    \item The KL-loss $L^P_{KL}$ for the regularity of $\vb{m}^P$, as the KL-divergence between $N({\mu}^P, {\Sigma}^P)$ and the standard normal distribution. 
\end{itemize}
The overall loss for $\vb{P}$ sums them up: $L_P=\lambda_1 L_{comp}+\lambda_2 L_{trj}+\lambda_3 L^P_{KL}$.

\subsubsection{A-Block} models the relationship between the mid-level feature and action: it exploits the long-term time span to aggregate the sequence of mid-level features $\vb{m}^P$ from the whole observation, and predicts a sequence of mid-level features $\vb{m}^A$ for future timestamps, which are further expanded by $\vb{D}^P$ into concrete motion. 
In addition to variational auto-encoding, $\vb{A}$ has its latent bottleneck feature also aligned with text embeddings of the action taxonomy, to enable action recognition of the observation and action-controlled prediction.

The encoder $\vb{E}^A$ derives action from hand motion and object features across the observation. Its input sequence concatenates the trainable tokens $\tilde{\mu}^A, \tilde{\Sigma}^A\in\mathbb{R}^d$ with the clip-wise mid-level $\vb{m}^P_{1:n}$ and object feature ${\vb{\omega}}_{1:n}$ ($n=\lceil T/t\rceil$). 
Specifically, $\vb{m}^P_i$ is the $\mu^P$ from $\vb{E}^P$; $\vb{\omega}_i$ is comparable to the CLIP~\cite{cherti2023reproducible, Radford2021LearningTV,schuhmann2022laionb} feature of object name, which is aggregated by an extra individual $\vb{E}^{PO}$ from the per-frame object features (\cref{sec:method_training}).
We further add a sinusoidal phase encoding $\phi_i$ to $\vb{m}^P_i$ and $\vb{\omega}_i$, which denotes the number of clips since the beginning of the performed action. 
Given $\mu^A,\Sigma^A$ output from $\vb{E}^A$, we follow $N(\mu^A,\Sigma^A)$ to re-parameterize and obtain the bottleneck latent feature $\vb{\alpha}$.

We then inject $\vb{\alpha}$ into the decoder $\vb{D}^A$ to enable action-controlled generation. For the cross attention of $\vb{D}^A$, we utilize $\vb{\alpha}$ and optionally include the clip-wise feature obtained from $\vb{E}^A$ for enhanced continuity. 
The parallel decoder $\vb{D}^A$ takes the phase embeddings $\phi_{n+1:n+\bar{n}}$ as input, and outputs $\vb{m}^A_{n+1:n+\bar{n}}$ depicting the mid-level features of the future $\bar{n}$ consecutive clips. 
One can further expand the predicted $\vb{m}^A$ into concrete poses through $\vb{D}^P$, completing the cycle of long-term observation for long-term prediction, with consistency in both global action and local poses (\cref{sec:method_tasks}, P.b). 

To train $\vb{A}$, besides the KL-loss $L_{KL}^A$, we constrain the bottleneck latent $\vb{\alpha}$ by matching it with the embedding of action taxonomy $\mathcal{A}=\{a=\mathtt{FC_1}(\overline{\vb{\alpha}})\in\mathbb{R}^d\}$, where $\overline{\vb{\alpha}}$ are CLIP text embeddings~\cite{cherti2023reproducible,Radford2021LearningTV,schuhmann2022laionb} for action labels in the taxonomy of size $N_A$. 
Therefore, the action recognition loss is:

\begin{equation}
L_{action}=\sum_{i=1}^{N_A} w_i\left(||\vb{\alpha}-a_i||_1 - \log\text{Pr}(a_i|\vb{\alpha})\right)\label{eq:action}
\end{equation}
which penalizes the differences of action features by both $l_1$-norm and contrastive similarity.
Here, $w_i$ is 1 for the GT action and 0 otherwise, and

\begin{equation}
\text{Pr}(a_i|\vb{\alpha})=\frac{\text{exp}(\hat{\vb{\alpha}}\cdot\hat{a}_i/\tau)}{\sum_{j=1}^{N_A}\text{exp}(\hat{\vb{\alpha}}\cdot\hat{a}_j/\tau)}
\label{eq:action_prob}
\end{equation}
measures the probabilistic similarity of predicted and GT labels among candidates from taxonomy.
$\hat{z}=z/||z||$ denotes the normalized unit vector, and $\tau=0.07$ is the temperature of contrastive similarity.
When testing, we perform action recognition by searching the closest labels to $\mu^A$.

For future motion supervision, instead of expanding down to concrete hand poses, we directly compare mid-level features for efficiency. 
Specifically, $\vb{m}^A_{n+1:n+\bar{n}}$ are compared with pre-computed $\overline{\vb{m}}^P_{n+1:n+\bar{n}}$, with $\overline{\vb{m}}^P_j = \mu^P_j$ encoding the GT hand motion of the future $j$-th clip via $\vb{E}^P$.
The motion prediction loss is

\begin{equation}
L_{mid}=\sum_{j=n+1}^{n+\bar{n}}||\vb{m}^A_j-\overline{\vb{m}}^P_j||_1
\end{equation} 

To summarize, the overall loss of $\vb{A}$ is $L_A=\lambda_4 L_{mid}+\lambda_5 L_{action}+\lambda_6 L_{KL}^A$.

\subsection{Network Flow for Tasks}\label{sec:method_tasks}

The framework addresses tasks of recognition (\ie~pose refinement and action recognition) and prediction by going through different paths within the network.

\noindent\textbf{Recognition} tasks are performed through the encoders. Specifically, $\vb{E}^P$ refines the input per-frame estimated hand pose by referring to the motion regularity over a subsecond clip of $t$ frames, followed by $\vb{E}^A$ to output $\mu^A$ for action recognition over the entire $T$ input frames.

\noindent\textbf{Prediction} of future hand motion is fulfilled by the decoders. 
Given observed hand motion $\tilde{\vb{H}}_{1:T}$, G-HTT provides two ways for the prediction of diverse and realistic hand motions (\cref{fig:hierarchy}): 
\begin{itemize}[leftmargin=*,labelsep=2em, align=parleft]
    \item[(P.a)] \textbf{(1)$\to$(4)}. 
    It generates locally consistent motions using only $\vb{P}$. It takes the last $t$ observed frames $\tilde{\vb{H}}_{T-t+1:T}$ as input, and predicts motion of the following clip $\vb{H}_{T+1:T+t}$ by sampling from $N(\mu^P,\Sigma^P)$ for $\vb{D}^P$ decoding. 
    The output motion can then be autoregressively fed back to $\vb{P}$ as input for longer prediction.

    \item[(P.b)] \textbf{(1)$\to$(2)$\to$(3)$\to$(4)}. For more realistic long-term prediction with action guidance, we move up the hierarchy to leverage $\vb{A}$ and predict $\vb{m}^A_{n+1:n+\bar{n}}$, with diversity coming from sampling from $N(\mu^A,\Sigma^A)$.
    $\vb{m}^A_{n+1:n+\bar{n}}$ are further decoded by $\vb{D}^P$ into concrete poses $\vb{H}_{T+t+1:T+(\bar{n}+1)t}$. 
\end{itemize}
We compare the two paths for motion prediction empirically in \cref{sec:exp_hierarchy}, and use P.b by default for long-term prediction in other experiments.

\subsection{Implementation Details}\label{sec:method_training}
$\vb{P}$, $\vb{A}$ are trained across different datasets separately.
To deploy G-HTT for practical RGB video processing, we leverage an external image-based hand object estimator $\vb{F}$ and a sequence-based object aggregator $\vb{E}^{PO}$, to provide the input for G-HTT.
The external modules can be trained independently or obtained from off-the-shelf models.

\subsubsection{G-HTT Details} We set $t=16$ and allow $T$ to have a maximum value of 256 at 30 fps.
Both $\vb{P}$ and $\vb{A}$ have 9 layers for the encoders and decoders, with a token dimension of $d=512$. We train a single network across datasets with different pose and action annotation qualities for enhanced capability (\cref{sec:exp_others}).

We first train $\vb{P}$ on all available pose sequences, regardless of the availability and transition of action labels, thanks to the decoupling of action and $\vb{P}$.
We augment the input motion $\tilde{\vb{H}}$ with random noise, making $\vb{P}$ capable of coping with noisy per-frame estimation in the recognition stage. Then, we fix the pre-trained $\vb{P}$ and train $\vb{A}$, whose training data assumes the same action shared between the observation and prediction. 
We derive the mid-level $\overline{\vb{m}}^P$ from the pose annotations with $\vb{E}^P$, randomly divide the training sequence into observed and predicted parts, and correspondingly assign $\overline{\vb{m}}^P$ as the input of $\vb{E}^A$ or supervision signal of $\vb{D}^A$. 
For the input of $\vb{E}^A$, we augment the mid-level features with random Gaussian noise, and refer to the noun of GT action for the clip-wise object feature $\vb{\omega}$.

We use AdamW~\cite{loshchilov2017decoupled}
optimizer with a learning rate of $10^{-4}$ and weight decay of 0.01 for both $\vb{P}$ and $\vb{A}$, where our batch size is 256. We respectively train $\vb{P}$ and $\vb{A}$ with 80 and 200 epochs, with loss weights as $\lambda_1,\lambda_2=1, \lambda_3=10^{-5}$, and $\lambda_4=1, \lambda_5=0.1, \lambda_6=10^{-5}$. \supp{Other design and training details are illustrated in the supplementary.}

\subsubsection{Image-based Estimator} 
$\vb{F}$ takes an image as input, and outputs for the image its hand pose, along with the object feature $\tilde{\vb{o}}$ that is comparable with CLIP~\cite{cherti2023reproducible,Radford2021LearningTV,schuhmann2022laionb} feature of the object in manipulation.
For experiments, we implement $\vb{F}$ as a ResNet-18~\cite{he2016deep} backbone followed by heads regressing the hand pose and object feature.
\supp{We provide more details in the supplementary.}

\subsubsection{Clip-wise Object Detector}
$\vb{E}^{PO}$ extracts $\vb{\omega}$ as the clip-wise object representation aggregated from per-frame object features $\tilde{\vb{o}}$ of $t$ consecutive frames. The clip-wise $\vb{\omega}$ is then fed into $\vb{E}^A$ to provide a consistent object information for action recognition (\cref{sec:method_hierahrcy}). 
We implement $\vb{E}^{PO}$ by 2 transformer encoder layers, which is trained based on $\vb{F}$; \supp{details are explained in the supplementary.}

\section{Experiments}
\subsection{Datasets}\label{sec:dataset}

We use three hand action datasets \cite{kwon2021h2o,sener2022assembly101,ohkawa2023assemblyhands} covering highly diverse motions and actions for training and testing. Across all datasets, we consider $N=20$ joints annotated by~\cite{kwon2021h2o}, leaving the carpometacarpal joint of the thumb out as its annotation is unavailable in~\cite{sener2022assembly101}.
As short-span hand motions are independent of action, we train $\vb{P}$ on untrimmed sequences that could contain multiple action annotations, and train $\vb{A}$ on trimmed sequences with clean and complete action annotations. 
We use trimmed sequences with clean action labels to evaluate both pose and action.

\noindent\textbf{H2O~\cite{kwon2021h2o}} records four subjects performing 36 indoor daily activities, in four fixed (\textit{cam0-3}) and one egocentric camera (\textit{cam4}) viewpoints. 
We conduct evaluation on validation and test splits, the latter having subjects unseen in training.

\noindent\textbf{Assembly101~\cite{sener2022assembly101} and AssemblyHands~\cite{ohkawa2023assemblyhands}} 
Assembly101~\cite{sener2022assembly101} contains procedures of people assembling and disassembling toy vehicles, with pose labels computed automatically by UmeTrack~\cite{han2022umetrack}. AssemblyHands~\cite{ohkawa2023assemblyhands} further improves the pose annotation quality for a subset of Assembly101 sequences. 
We follow the splits of Assembly101 for training and testing, using actions of its fine-grained taxonomy with 1380 different labels. 
We conduct evaluations on sequences that have more reliable pose annotations from AssemblyHands, \supp{and consider six fixed camera views that have no severe hand occlusions (details in supplementary). }
We focus evaluation on the validation split, as it contains accessible object labels for action recognition and motion prediction (\cref{sec:exp_jointrp,sec:exp_hierarchy}), which the test split lacks.

\subsection{Setup and Metrics}\label{sec:metrics}
\noindent\textbf{Recognition} 
We take a whole video sequence depicting the process of an action as input. 
For pose estimation, we use metrics of \textbf{MPJPE-RA} (Mean Per Joint Position Error-Root Aligned) and \textbf{MPJPE-PA} (Procrustes Analysis of MPJPE) (unit: \textit{mm}). 
To deal with the ambiguity of different hand scales, we align estimation and GT by equalizing the average length of palm bones. 
For action recognition, we report the top-1 classification accuracy (\textbf{Action Acc.}).

\noindent\textbf{Prediction}  
We divide each action video into segments of 16 frames. 
Given each segment, we use its pose and object annotation as the input observation, and predict the rest sequence until reaching the end of action or the maximum duration of 96 frames.
We generate 20 random samples from $N(\mu,5\Sigma)$ as a trade-off between accuracy and diversity (\cref{sec:exp_jointrp,sec:exp_hierarchy}). 
For the evaluation of generative results, we mainly use the widely adopted \textbf{FID} (Frechet Inception Distance) \cite{FID17} to assess quality, which computes the distributional distance of features between generated and GT motion sequences. 
The features are obtained from the last layer of a pre-trained transformer-based action recognition network, and the GT sequences are from the evaluation split unseen in training. 
A smaller FID means a more faithful generation.
In addition, to explicitly measure generation diversity, we report \textbf{APD} (Average Pairwise Diversity)~\cite{yuan2020dlow} (unit: \textit{mm}) that computes the average distance between all pairs of 20 generated samples. 
A larger APD means a more diverse generation.
\supp{More details about the setup and metrics are given in the supplementary.}

\subsection{Joint Modeling of Recognition and Prediction}\label{sec:exp_jointrp}

We first demonstrate the enhanced capability because of our joint modeling of recognition and prediction, by respectively comparing with state-of-the-art solutions for either recognition (\ie HTT~\cite{wen2023hierarchical}) or prediction (\ie PoseGPT~\cite{lucas2022posegpt}). \supp{More implementation details for the baselines are given in the supplementary.}

\begin{table}[!t]
\centering
\caption{Pose estimation and action recognition results. For hand pose estimation we report MPJPE-RA/-PA in $mm$ for (left,right) hand respectively, 
where $*$ denotes training views leveraged for HTT~\cite{wen2023hierarchical}. \supp{Please refer to the supplementary for complete results on all camera views, and comparison on the H2O-Val}.}\label{tab:recognition}
\begin{subtable}[t]{0.49\textwidth}
\centering
\resizebox{0.99\linewidth}{!}{
\begin{tabular}{C{1cm}|C{2cm}|C{2cm}|C{2cm}|C{2cm}}
\hline 
\multicolumn{2}{c|}{H2O-Test} & Resnet-18($\vb{F}$) & HTT~\cite{wen2023hierarchical} & Ours \\
\hline
\multirow{3}{*}{cam0$^*$} & MPJPE-RA$\downarrow$ & 27.0,25.6 &26.9,\textbf{24.1} & \textbf{26.5},25.3 \\
& MPJPE-PA$\downarrow$ & 7.8,10.6 & \textbf{7.3},10.4&7.4,\textbf{10.3}\\ 
& Action Acc.$\uparrow$ & - & \textbf{85.12}&59.92\\
\hline

\multirow{3}{*}{cam2} & MPJPE-RA$\downarrow$& 19.2,24.8&20.1,25.4&\textbf{18.9,24.5}\\
& MPJPE-PA$\downarrow$&6.9,10.6&7.4,11.0&\textbf{6.6,10.3}\\
& Action Acc.$\uparrow$&-&\textbf{73.55}&68.18\\
\hline

\multirow{3}{*}{cam4} & MPJPE-RA$\downarrow$ & 18.4,21.4&101.2,137.8&\textbf{17.9,21.0}\\
& MPJPE-PA$\downarrow$& 6.8,9,4&28.5,33.8&\textbf{6.4,9.1} \\
& Action Acc.$\uparrow$& -&2.89&\textbf{57.85}\\
\hline
\end{tabular}}
\end{subtable}
\begin{subtable}[t]{0.49\textwidth}
\centering
\resizebox{0.99\linewidth}{!}{
\begin{tabular}{C{1cm}|C{2cm}|C{2cm}|C{2cm}|C{2cm}}
\hline
\multicolumn{2}{c|}{AssemblyHands-Val} & Resnet-18($\vb{F}$) & HTT~\cite{wen2023hierarchical} & Ours \\
\hline
\multirow{3}{*}{v1} & MPJPE-RA$\downarrow$& 35.4,22.7& 55.6,39.0&\textbf{35.1,22.4}\\
& MPJPE-PA$\downarrow$& 12.0,10.8 & 17.2,14.2& \textbf{11.7,10.4}\\
& Action Acc.$\uparrow$&-&16.55&\textbf{36.01}\\
\hline
\multirow{3}{*}{v3$^*$} & MPJPE-RA$\downarrow$&27.5,27.2&\textbf{26.7},27.3&27.3,\textbf{26.9}\\
& MPJPE-PA$\downarrow$&12.2,12.0&12.3,12.1&\textbf{11.9,11.7}\\
& Action Acc.$\uparrow$&-&\textbf{39.42}&34.79\\
\hline
\multirow{3}{*}{v8} & MPJPE-RA$\downarrow$ & 26.1,30.4&91.3,88.5&\textbf{25.9,30.0}\\
& MPJPE-PA$\downarrow$&11.8,12.3&24.2,27.3&\textbf{11.5,11.8}\\
& Action Acc.$\uparrow$ & - &9.98&\textbf{36.74}\\
\hline
\end{tabular}}
\end{subtable}
\end{table}

\begin{figure*}[!t]
\centering
\includegraphics[width=0.9\linewidth]{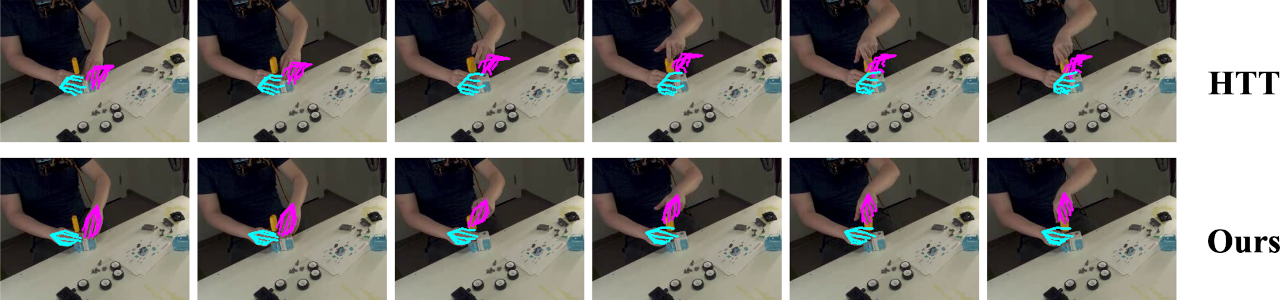}
\caption{Qualitative comparison of pose estimation for HTT~\cite{wen2023hierarchical} and ours, on camera view \textit{v1} of Assembly datasets~\cite{ohkawa2023assemblyhands,sener2022assembly101}. \supp{More cases are provided in the supplementary.}}\label{fig:quali_recognition}
\end{figure*}

\subsubsection{Recognition} The most relevant baseline is HTT~\cite{wen2023hierarchical}, which also models the semantic-temporal hierarchy but focuses only on recognition. 
Based on the pre-trained image-based estimator $\vb{F}$ used by ours for fair comparison (\cref{sec:method_training}), we train HTT on two camera views of H2O (\textit{cam0,1}) and one view (\textit{v3}) of AssemblyHands, where we concatenate image feature with the estimated hand pose and object from $\vb{F}$ as the per-frame input of HTT. 
We also take the initial pose estimation of $\vb{F}$ as a reference for comparison. 
Moreover, to obtain the object input for both methods, on H2O we leverage the network estimation, while on AssemblyHands we use the GT labels, where it is very challenging to recognize objects reliably due to cluttered scenes and frequent occlusions (\cref{fig:quali_recognition}).

As shown in \cref{tab:recognition} and \cref{fig:quali_recognition}, G-HTT demonstrates robust accuracy on various camera views, for refining the local pose and action recognition, even though G-HTT is never trained on $\vb{F}$. 
In comparison, although HTT fits better on views that are trained on or close to trained ones (\textit{e.g.,} \textit{cam2} of H2O), its performance significantly degrades on the other views, even worse than its input obtained from $\vb{F}$.
The results show that our simultaneous modeling of both recognition and prediction enhances generalization by learning regular motion priors across tasks. 
In contrast, a recognition-only network is more likely to overfit particular data distributions.

\begin{figure*}[!t]
\centering
\includegraphics[width=0.99\linewidth]{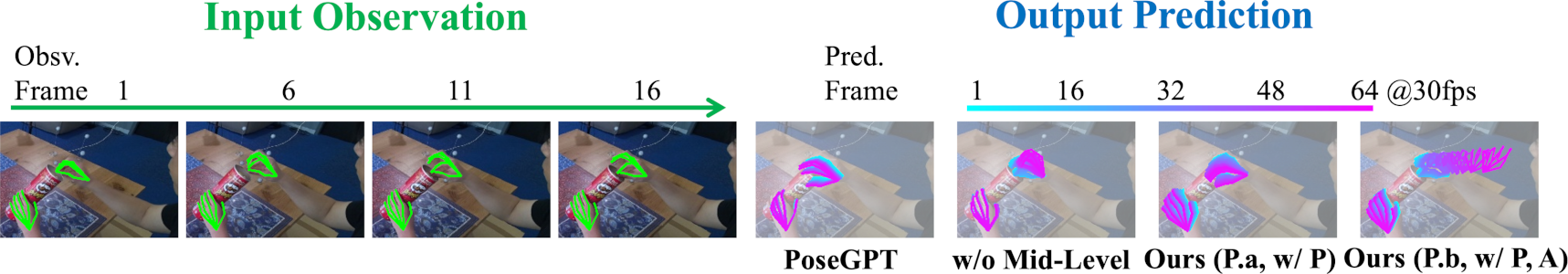}
\caption{Qualitative comparison of predicted motions for PoseGPT~\cite{lucas2022posegpt},  the ablated settings of w/o mid-level, w/ only $\vb{P}$ via path P.a, and the full G-HTT (w/ $\vb{P},\vb{A}$, via path P.b) on H2O. 
\supp{More qualitative cases are provided in the supplementary.}}\label{fig:quali_prediction}
\end{figure*}

\begin{table}[!t]
\centering
\caption{Comparison with PoseGPT~\cite{lucas2022posegpt} for motion prediction, on action sequences that are longer than 1 sec. APD  in \textit{mm} for (left, right) hand respectively.}\label{tab:prediction_posegpt}
\resizebox{0.9\linewidth}{!}{
\begin{tabular}{c|c|C{1.5cm}C{1.5cm}|C{1.5cm}C{1.5cm}|C{1.5cm}C{1.5cm}}
\hline 
\multirow{2}{*}{}& \multirow{2}{*}{GT action input} & \multicolumn{2}{c|}{H2O-val} & \multicolumn{2}{c|}{H2O-Test} & \multicolumn{2}{c}{AssemblyHands-Val} \\
&&  FID$\downarrow$ &  APD$\uparrow$ & FID$\downarrow$ & APD$\uparrow$ & FID$\downarrow$ &  APD$\uparrow$ \\
\hline
PoseGPT~\cite{lucas2022posegpt} & $\checkmark$ & \textbf{5.19} & \textbf{32.3,43.1} & 11.70 & \textbf{24.1,48.6} & 16.07 & 25.3,\textbf{33.0}\\
Ours (P.b, w/ $\vb{P},\vb{A}$) & $\times$ &  5.32  & 22.1,25.7 & \textbf{8.19}  & 20.1,33.9 & \textbf{5.04} & \textbf{28.1},32.8 \\
\hline
\end{tabular}}
\end{table}

\subsubsection{Prediction} 
We take PoseGPT~\cite{lucas2022posegpt}, a state-of-the-art model for motion prediction with \textit{prescribed} action, as a baseline for performance evaluation. 
While the original PoseGPT trains on body motion, we retrain PoseGPT with its official code by combining the three hand pose-action datasets of Assembly101, AssemblyHands and H2O as we do.
We evaluate on action sequences longer than 1 sec to better show the differences in prediction.

As reported in \cref{tab:prediction_posegpt}, G-HTT shows significantly better FID on the H2O-test split of unseen subjects and on AssemblyHands; meanwhile, the two methods have comparable accuracy on the H2O-val split of trained subjects. 
These results show our better generation quality across actions and datasets.
Visually from \cref{fig:quali_prediction}, while PoseGPT suffers from lacking regularity for predicted motion, our prediction shows globally consistent action.

We attribute the differences to two factors: our joint learning of both recognition and prediction, and our hierarchical model for pose and action.
In contrast to action-conditioned generation alone, by modeling both recognition and prediction our framework learns strong motion-action regularities across datasets covering highly diverse hand actions, as shown by the consistent high quality across three datasets (\cref{tab:prediction_posegpt}).
Moreover, different from the vector quantization of PoseGPT for clip-wise motion, our mid-level representation originates from $\vb{P}$, where $\vb{P}$ models not only the observed motion but also prediction, enabling global action-guided motion generation that preserves local motion continuity (see also \cref{sec:exp_hierarchy}).

\subsection{Modeling Semantic-Temporal Hierarchy}\label{sec:exp_hierarchy}
In this section, we examine the effects of modeling the semantic-temporal hierarchy. 
As HTT~\cite{wen2023hierarchical} has well demonstrated the benefits of leveraging this hierarchy in recognition tasks, here we mainly examine its benefits for motion prediction, especially on sequences longer than 2 secs where global action is more apparent.

\begin{table}[!t]
\centering
\caption{Comparison of long-term motion prediction decoded from $\vb{m}^P$ (P.a) and $\vb{m}^A$ (P.b), on action sequences that are longer than 2 sec.}\label{tab:prediction_ponly}
\resizebox{0.8\linewidth}{!}{
\begin{tabular}{c|C{1.5cm}C{1.5cm}|C{1.5cm}C{1.5cm}|C{1.5cm}C{1.5cm}}
\hline 
\multirow{2}{*}{}&  \multicolumn{2}{c|}{H2O-val} & \multicolumn{2}{c|}{H2O-Test} & \multicolumn{2}{c}{AssemblyHands-Val} \\
&  FID$\downarrow$ &  APD$\uparrow$ & FID$\downarrow$ & APD$\uparrow$ & FID$\downarrow$ &  APD$\uparrow$ \\
\hline
Ours (P.a, w/ $\vb{P}$) & 8.18 &  \textbf{40.7,48.3} & 12.78 & \textbf{36.7,52.5} & 8.20 &  \textbf{40.8,51.1}\\
Ours (P.b, w/ $\vb{P},\vb{A}$)  & \textbf{6.59}  & 29.7,33.3 & \textbf{10.88} & 26.2,48.3 & \textbf{6.84} & 34.8,42.3 \\
\hline
\end{tabular}}
\end{table}

\subsubsection{Action for Prediction} We first compare the long-term prediction decoded from $\vb{m}^P$ and $\vb{m}^A$, \ie, the two strategies P.a, P.b described in \cref{sec:method_tasks}, to examine the effectiveness of involving $\vb{A}$ for long-term prediction. 
As shown in \cref{tab:prediction_ponly} and \cref{fig:quali_prediction}, generations from $\vb{m}^A$ are more realistic and plausible, with lower FID and more consistent global motion. 
In contrast, results of P.a show larger diversity due to its short-term modeling, but lack fidelity or regularity for long-term motion.
Overall, the comparison shows the importance of action modeling in generating faithful and action-guided motions.

\begin{table}[!t]
\centering
\caption{Comparison of motion prediction between ours and the ablated setup without modeling the mid-level, on action sequences that are longer than 2 sec.}\label{tab:prediction_flatten}
\resizebox{0.9\linewidth}{!}{
\begin{tabular}{c|c|C{1.5cm}C{1.5cm}|C{1.5cm}C{1.5cm}|C{1.5cm}C{1.5cm}}
\hline 
\multirow{2}{*}{}&  \multirow{2}{*}{Sampling with}&  \multicolumn{2}{c|}{H2O-val} & \multicolumn{2}{c|}{H2O-Test} & \multicolumn{2}{c}{AssemblyHands-Val} \\
& &  FID$\downarrow$ &  APD$\uparrow$ & FID$\downarrow$ & APD$\uparrow$ & FID$\downarrow$ &  APD$\uparrow$ \\
\hline
w/o Mid-Level & $\mu,5\sigma$ & \textbf{6.69} & 18.0,23.7 & 13.84 &  17.9,29.4 &  5.67 &16.9,21.9\\
w/o Mid-Level & $\mu,10\sigma$ & 14.05 & 25.5,\textbf{33.4} & 22.64 &  24.1,35.2 & 6.31 & 23.3,30.2\\
Ours (P.b, w/ $\vb{P},\vb{A}$)  & $\mu,5\sigma$ & 9.30  & \textbf{27.8},31.4  & \textbf{13.10} & \textbf{24.6,43.3}  & \textbf{5.21} &\textbf{33.6,40.0}\\
\hline
\end{tabular}}
\end{table}

\subsubsection{Mid-level for Prediction} 
In addition to enabling a decoupled training strategy for $\vb{P}$ and $\vb{A}$ (\cref{sec:method_training}), the modeling of mid-level features should enhance the learning of generation.
To verify it, we construct a flattened baseline (\ie, w/o mid-level) by removing the mid-level representation and instead using a single transformer VAE to directly model pose and action.
For this flattened baseline, its encoder takes hand poses and objects as input and outputs for action recognition; its decoder directly outputs the future hand poses.
The flattened baseline has a comparable amount of parameters as $\vb{P}$ and $\vb{A}$, and is trained on the same dataset as our framework for fair comparison.

From~\cref{tab:prediction_flatten} we find that under a comparable FID, the mid-level representation enables better diversity (3rd, 5th rows); meanwhile, as we increase the generation diversity of the flattened baseline via more noisy sampling, its accuracy significantly degrades (4th, 5th rows).
From \cref{fig:quali_prediction}, we can see that the flattened baseline results lack global regularity, despite its modeling of pose and action through a powerful end-to-end transformer VAE.
The comparison shows that the mid-level representation enables easier learning of global motion regularity, as it decouples the complex task of action-guided motion generation into hierarchical subtasks better captured by $\vb{P}$ and $\vb{A}$ respectively.

\subsection{Discussion}\label{sec:exp_others}
We make more observations about motion prediction, training strategy, and mid-level representation, to give additional understanding of the framework. 
Due to space limit, \supp{we provide more details in supplementary}.

\noindent\textbf{Training with Assembly101} 
We find that including the large-scale Assembly101 for training G-HTT significantly benefits motion prediction on H2O and action recognition, although the pose annotation of Assembly101 is not sufficiently accurate for training pose estimators (\cf~\cite{ohkawa2023assemblyhands}).
The finding points to the importance of large-scale pretraining of fundamental modules.

\noindent\textbf{Predicting Action Transition} 
We observe that our model can generate smooth transitions between actions, (\eg motion depicting \textit{place spray} $\to$ \textit{apply spray}), although this is never explicitly trained for.
It probably comes from training $\vb{P}$ on sequences of mixed action annotations (\cref{sec:method_training}), which allows $\vb{P}$ to drive the transition by decoding local motions into new actions.
Meanwhile, it is also facilitated by the decoupled training of $\vb{P}$ and $\vb{A}$.

\noindent\textbf{The Mid-level Regularity} 
We blend mid-level features of two different input pose sequences, and decode the blended mid-level features into pose sequences.
The generated motions naturally interpolate between the tendencies of two given inputs, showing the regularity of the learned mid-level representation.

\section{Conclusion}

We present a novel unified framework for understanding hand pose and action, which concurrently models both recognition and prediction, and captures the hierarchy of semantic dependency and temporal granularity.
The framework addresses tasks of 3D hand pose refinement, action recognition, and 3D hand motion prediction, showing improved performances than isolated solutions.
The framework has two cascaded Transformer VAE blocks to model pose and action respectively. Both blocks have their encoder and decoder output respectively for recognition and prediction, while their VAE bottleneck extracts the temporal regularity synergized between the two sides.
A mid-level clip-wise motion representation is further introduced to bridge the two blocks. 
The connected cascade enables regular pose and action modeling over both short and long time spans, and brings flexibility to train the two blocks separately on multiple datasets with different setups and annotation granularities.
Extensive experiments validate the performance and design of our framework on both recognition and prediction across different datasets.

\noindent\textbf{Limitations and Future Work} 
We assume a fixed camera viewpoint for input videos; to process cases with drastic camera movement (\eg, egocentric views with large head motions), an explicit decomposition of hand and camera motion would be necessary. 
Another aspect is to leverage hand motion as priors for robust recognition of manipulated objects, therefore further benefiting action understanding. 
Moreover, extensions to cross-dataset settings and human body pose action modeling are also interesting directions for broader impacts.

\noindent\textbf{Acknowledgement} 
This research is supported by the Innovation and Technology Commission of the HKSAR Government under the InnoHK initiative, Innovation and Technology Commission (Ref: ITS/319/21FP), Research Grant Council of Hong Kong (Ref: 17210222, 17200924), JST ASPIRE (Grant Number: JPMJAP2303), and JST ACT-X (Grant Number: JPMJAX2007).

\clearpage
\setcounter{page}{1}

\appendix
\section{Overview}
The supplementary document is organized with the following sections, to support the main text with setup details and additional results:

\begin{itemize}
    \item \noindent\cref{sec:supp_implementation} elaborates the detailed implementation of our network blocks.

    \item \noindent\cref{sec:supp_eval_setup} illustrates additional details for the evaluation setup and baseline implementation.

    \item \noindent\cref{sec:supp_recognition} provides additional results for the recognition tasks.

    \item \noindent\cref{sec:supp_prediction} shows additional qualitative results for the motion prediction.

    \item \noindent\cref{sec:supp_discussion} supplements the discussion for~\cref{sec:exp_others} of main text.
    
\end{itemize}

We also provide a supplementary video, to show qualitative results mentioned in~\cref{sec:supp_recognition,sec:supp_prediction,sec:supp_discussion}.

\section{Implementation Details}\label{sec:supp_implementation}
\subsection{Notations}

In~\cref{tab:supp_notation}, we summarize the notations and their meanings used in the main text.

\subsection{3D Hand Motion Representation}
Given a sequence depicting two interacting hands, we construct the 3D hand representation for each frame as 
$\vb{H}=(p^L,p^R,v^L,v^R,r)$, where $p^L,p^R\in \mathbb{R}^{3N}$ are the 3D coordinates for $N$ joints of each hand aligned to a predefined template~\cite{javier2017mano}, $v^L,v^R\in \mathbb{R}^{9}$ are the relative rigid palm transformations compared with the previous frame, and $r\in\mathbb{R}^{9}$ are the relative rigid palm transformation between two hands in the current frame. We denote the rigid transformation $v^L,v^R,r$ by concatenating the 6D continuous representation of rotation~\cite{zhou2019continuity} and 3D translation.

\begin{table*}[!tp]
\centering
\caption{Summary of important notations.}\label{tab:supp_notation}
\renewcommand\arraystretch{1.2}
\resizebox{0.99\linewidth}{!}{
\begin{tabular}{c|ccc}
\hline 
&Symbol &Dimension & Meaning \\
\hline
\multirow{6}{*}{\tabincell{c}{Constants}}&$d$ & - & token dimension of $\vb{P},\vb{A},\vb{E}^{PO}$\\
&$N$ & - & number of hand joints for each of the left and right hand \\
&$T$ & - & number of consecutive frames for inputs of G-HTT\\
&$t$ & - & number of consecutive frames for a clip-wise time span, as leveraged by $\vb{E}^P,\vb{D}^P,\vb{E}^{PO}$\\
&$n$ & - & number of clips for inputs of $\vb{E}^A$, with $n=\lceil T/t\rceil$ \\
&$\overline{n}$ & - & number of clips for outputs of $\vb{D}^A$ \\

\hline
\multirow{6}{*}{\tabincell{c}{Blocks}}&$\vb{P}$ &-& pose block \\
&$\vb{E}^P,\vb{D}^P$ &-& encoder and decoder of $\vb{P}$ \\
&$\vb{A}$ &-& action block \\
&$\vb{E}^A,\vb{D}^A$ &-& encoder and decoder  of $\vb{A}$ \\
&$\vb{F}$ &-& image-based estimator for frame-wise hand pose and object feature\\
&$\vb{E}^{PO}$ &-& clip-wise object detector\\
\hline

\multirow{5}{*}{\tabincell{c}{Frame-wise\\3D hand\\representation}}&$p^L$ & $\mathbb{R}^{3N}$ & hand joint coordinates of the left hand aligned to predefined template~\cite{javier2017mano}\\
&$p^R$ & $\mathbb{R}^{3N}$  & hand joint coordinates of the right hand aligned to predefined template~\cite{javier2017mano}\\
&$v^L$ & $\mathbb{R}^{9}$  &relative rigid left-hand palm transformation compared with previous frame \\
&$v^R$ & $\mathbb{R}^{9}$  &relative rigid right-hand palm transformation compared with previous frame \\
&$r$ & $\mathbb{R}^{9}$ &relative rigid palm transformation between left and right hands in the current frame \\
\hline

\multirow{2}{*}{\tabincell{c}{Inputs\\of $\vb{P}$}}& $\tilde{\mu}^P,\tilde{\Sigma}^P$ & $\mathbb{R}^{d}$  & input trainable token for $\vb{E}^P$ \\
& $\tilde{\vb{H}}$ & $\mathbb{R}^{147}$  & input frame-wise 3D hand representation for  $\vb{E}^P$\\

\hline

\multirow{5}{*}{\tabincell{c}{Outputs\\of $\vb{P}$}} & $\mu^P,\Sigma^P$ & $\mathbb{R}^{d}$ & output tokens of $\vb{E}^P$, to re-parameterize the normal distribution for sampling $\vb{m}^P$\\
&$\vb{m}^P$ & $\mathbb{R}^{d}$ & mid-level feature as the latent bottleneck of $\vb{P}$\\
&$\vb{H}$ & $\mathbb{R}^{147}$ & output frame-wise 3D hand representation of $\vb{E}^P,\vb{D}^P$ \\
&$s^L$ & $\mathbb{R}^9$ & output frame-wise global pose of left-hand, obtained by accumulating $v^L$ of $\vb{H}$ outputted by $\vb{D}^P$\\
&$s^R$ & $\mathbb{R}^9$ & output frame-wise global pose of right-hand, obtained by accumulating $v^R$ of $\vb{H}$ outputted by $\vb{D}^P$\\

\hline

\multirow{3}{*}{\tabincell{c}{Supervision\\ signals\\ of $\vb{P}$}}&$\overline{\vb{H}}$ & $\mathbb{R}^{147}$ & GT frame-wise 3D hand representation \\
&$\overline{s}^L$ & $\mathbb{R}^9$ &  GT counterpart of $s^L$, obtained by accumulating $v^L$ of GT $\overline{\vb{H}}$\\
&$\overline{s}^R$ & $\mathbb{R}^9$ &  GT counterpart of $s^R$, obtained by accumulating $v^R$ of GT $\overline{\vb{H}}$\\
\hline

\multirow{4}{*}{\tabincell{c}{Inputs\\of $\vb{A}$}}&$\tilde{\mu}^A,\tilde{\Sigma}^A$ & $\mathbb{R}^{d}$  & input trainable token for $\vb{E}^A$ \\
&$\vb{m}^P$ & $\mathbb{R}^{d}$ & input clip-wise mid-level feature for $\vb{E}^A$\\
&$\omega$ & $\mathbb{R}^{d}$ & input clip-wise object feature for $\vb{E}^A$ \\
&$\phi$  & $\mathbb{R}^{d}$ & input clip-wise action phase feature for $\vb{E}^A,\vb{D}^{A}$\\
\hline

\multirow{4}{*}{\tabincell{c}{Outputs\\of $\vb{A}$}}&$\mu^A,\Sigma^A$ & $\mathbb{R}^{d}$ & output tokens of $\vb{E}^A$, to re-parameterize the normal distribution for sampling $\alpha$\\
&$\vb{\alpha}$ & $\mathbb{R}^{d}$ & global action feature as the latent bottleneck of $\vb{A}$\\

&$\eta_m$ & $\mathbb{R}^{d}$ & output clip-wise mid-level feature of $\vb{E}^A$ \\
&$\eta_o$ & $\mathbb{R}^{d}$ & output clip-wise object feature of $\vb{E}^A$ \\
&$\vb{m}^A$ & $\mathbb{R}^{d}$ & output clip-wise action-controlled mid-level feature of $\vb{D}^A$ \\

\hline
\multirow{3}{*}{\tabincell{c}{Supervision\\signals\\of $\vb{A}$}}&$\overline{\vb{\alpha}}$ & $\mathbb{R}^{d}$ & CLIP~\cite{cherti2023reproducible,Radford2021LearningTV,schuhmann2022laionb} action feature for actions in the taxonomy\\
&$a$ &  $\mathbb{R}^{d}$ & latent action feature to supervise $\vb{\alpha}$, obtained from $\overline{\vb{\alpha}}$ \\
&$\overline{\vb{m}}^P$ & $\mathbb{R}^{d}$ & clip-wise pre-computed mid-level feature to supervise $\vb{D}^A$, obtained by encoding GT $\overline{\vb{H}}$ with $\vb{E}^P$\\
\hline

\multirow{2}{*}{\tabincell{c}{$\vb{E}^{PO}$}} &
$\tilde{\vb{o}}$ &  $\mathbb{R}^{d}$ & input frame-wise object feature of $\vb{E}^{PO}$ \\
& $\omega$ &  $\mathbb{R}^{d}$ & output clip-wise object feature of $\vb{E}^{PO}$\\ 
\hline

\end{tabular}}
\end{table*}

\subsubsection{Compute $\vb{H}$ from Pose Trajectory} Given the pose sequence of $F$ frames for the left and right hand $h^L_{1:F},h^R_{1:F}$, with $h^L_i,h^R_i\in\mathbb{R}^{N\times 3},i\in[1,F]$ representing the 3D joint coordinates of each hand in a predefined global coordinate system, we follow the steps below to compute $\vb{H}_{1:F}$ from $h^L_{1:F},h^R_{1:F}$:

Given the predefined template $\vb{M}^L,\vb{M}^R$ for the left and right hand~\cite{javier2017mano}, we first respectively align the palm of $h^L_i, h^R_i$ with that of $\vb{M}^L,\vb{M}^R$ by adopting~\cite{umeyama1991least}. The corresponding rigid transformation is denoted as $a^L_i,a^R_i$, indicating the 3D rotation and translation to convert $h^L_i,h^R_i$ into the template space for obtaining $p^L_i,p^R_i$:
\begin{equation}
    p^L_i=\mathrm{W}(a^L_i, h^L_i), \ \ p^R_i= \mathrm{W}(a^R_i, h^R_i)
\end{equation}
with $\mathrm{W}(a,h)$ denoting transforming $h$ with $a$. We further normalize $p^L_i,p^R_i$ by referring to the hand scale computed as the average length of palm bones, and flatten the 3D coordinates to represent $p_i^L,p_i^R$ in $3N$-dimensional vectors.

Then, we compute $v_i^L, v_i^R$ from $a^L, a^R$. 
For $i>1$, we compute $v_i^L, v_i^R$ as 
\begin{equation}
v^{L}_i= (a^L_i)^{-1} \circ a^{L}_{i-1},\ \,v^{R}_i= (a^R_i)^{-1} \circ a^{R}_{i-1}
\end{equation}
with $\circ$ denoting the composition of different rigid transformations. For $i=1$, $v_i^L, v_i^R$ are assigned the identity transformation.

We derive $r_i$ from $(a^L_i)^{-1} \circ (a^R_i)$. We then concatenate $(p^L_i,p^R_i,v_i^L,v_i^R,r_i)$ as the frame-wise 3D hand representation.

\subsubsection{Recover Pose Trajectory from $\vb{H}$.} 
For $\vb{H}_{t+1:2t}$ outputted by $\vb{D}^P$, one can recover the pose trajectories $h^L_{t+1:2t},h^R_{t+1:2t}$ relative to $h^L_t,h^R_t$. 

To achieve this, for each frame $i\in[t+1,2t]$, we first obtain its relative rigid palm transformation $s^L_i,s^R_i\in\mathbb{R}^9$ compared with frame $t$, as computed by accumulating the $v^L_{t+1:i},v^R_{t+1:i}$:
\begin{equation}
s^L_{i}= v^L_{i}\circ v^L_{i-1}\circ \cdots \circ v^L_{t+1}, \ \  s^R_{i}= v^R_{i}\circ v^R_{i-1}\circ \cdots \circ v^R_{t+1}\label{eq:accum_trj}
\end{equation}    
with $\circ$ denoting the composition of rigid transformation. We further rescale $p^L_i,p^R_i$ by multiplying with given hand scales (\eg the average among the input sequence). Then, $h^L_i,h^R_i$ are recovered as
\begin{equation}\label{eq:local2base}
    h^L_i=\mathrm{W}(s^L_i, p^L_i), \ \ h^R_i=\mathrm{W}(s^R_i, p^R_i)
\end{equation}

For $\vb{H}_{1:t}$ outputted by $\vb{E}^P$, one can similarly recover the pose trajectories $h^L_{1:t},h^R_{1:t}$ relative to $h^L_1,h^R_1$. 
However, we notice that the accumulated error of $v^L,v^R$ can cause $h^L_{1:t},h^R_{1:t}$ to deviate from the image observations. Therefore, inspired by~\cite{starke2019neural}, we rectify the output trajectory by making $s^L_t,s^R_t$ consistent with $\tilde{s}^L_t,\tilde{s}^R_{t}$, where the latter are derived from the input $\tilde{\vb{H}}_{1:t}$. Specifically, we rectify $s^L_i,s^R_i,i\in[1,t]$ as:
\begin{equation}
\begin{split}
s^L_i\leftarrow s^L_i \circ \mathrm{L}(i, s^L_t, \tilde{s}^L_t)\\ 
s^R_i\leftarrow s^R_i \circ \mathrm{L}(i, s^R_t, \tilde{s}^R_t)
\end{split}
\end{equation}
where $\mathrm{L}(i,s_t,\tilde{s}_t)$ denotes linear interpolation between $s_t$ and $\tilde{s}_t$ with blending weights of $\lambda_1=(t-i)/(t-1)$ and $\lambda_2= 1-\lambda_1$, regarding the relative translation and rotation angle represented by $(s_t)^{-1} \circ \tilde{s}_t$. We then recover $h^L_{1:t},h^R_{1:t}$ via~\cref{eq:local2base} with the updated $s^L_{1:t},s^R_{1:t}$.

\subsection{Setup of G-HTT (\cref{sec:method_training} of Main Text)}
\subsubsection{Network}
Both $\vb{P}$ and $\vb{A}$ have 9 layers for the encoders and decoders, use the sinusoidal position encoding~\cite{vaswani2017attention} and GELU activation~\cite{hendrycks2016gaussian}. Each layer has the number of attention heads as 8, the token dimension as $d=512$, and the intermediate size of the feedforward layers as 2048; each layer first conducts layer normalization before the attention and feed-forward operations.

Furthermore, for $\vb{P}$, We encode the frame-wise input $\tilde{\vb{H}}_i$ into $\tilde{\gamma}_i$ before passing it into $\vb{E}^P$:
\begin{equation}
\tilde{\gamma}_i=[\mathtt{FC_2}([p^L_i,p^R_i]),\mathtt{FC_3}([v^L_i,v^R_i]),\mathtt{FC_4}(r_i)],
\end{equation}
where $[.]$ denotes feature concatenation, and $\mathtt{FC_2},\mathtt{FC_3},\mathtt{FC_4}$ respectively output $d/2$, $d/4$, $d/4$-dimensional features. 

And for output, we recover the frame-wise $\vb{H}_i=\mathtt{MLP_1}(\gamma_i)$ from the corresponding output token $\gamma_i$ of $\vb{E}^P$ or $\vb{D}^P$, 
with $\mathtt{MLP_1}$ having three layers of width $[d,d,6N+27]$ and adopting LeakyReLu activation for the hidden layers.

\subsubsection{Training Details for $\vb{P}$} We augment the input frame-wise $\tilde{\vb{H}}$ as described below:
\begin{itemize}
\item For $p^L,p^r$ normalized by hand scales, we augment by adding random noise from $N(\vec{\mathbf{0}},0.025\mathbf{I})$, with $\mathbf{I}$ denoting the identity matrix.
\item For $v^L,v^R$, the rotation is augmented by compositing with a random rotation which has an arbitrary axis and an angle sampled from $N(0,1.25^\circ)$. For the translation part, the added random noise follows $N(\vec{\mathbf{0}},0.75\mathbf{I})$ with a unit of $cm$. 
\item For $r$, we adopt a similar strategy as that for $v^L,v^R$. However, here the random rotation has its angle from $N(0,5^\circ)$, and the noise for translation follows $N(\vec{\mathbf{0}},5\mathbf{I})$ with a unit of $cm$. 
\end{itemize}

\subsubsection{Training Details for $\vb{A}$} We randomly divide the training sequence into observed and predicted parts, and set the maximum value of both $n$ and $\bar{n}$ to 16. To obtain the input $\vb{m}^P$ of $\vb{E}^A$, we add random noise sampled from $N(\vec{\mathbf{0}},1.5\mathbf{I})$ to the corresponding $\overline{\vb{m}}^P$.

\subsection{Setup of Image-based Estimator $\vb{F}$ 
(\cref{sec:method_training} of Main Text)}\label{sec:supp_ibe}

\subsubsection{Network} $\vb{F}$ takes as input an image ${I}\in\mathbb{R}^{3\times H\times W}$ with $H=270, W=480$, and output for $I$ its hand pose $\tilde{h}\in\mathbb{R}^{2N\times 3}$ and object feature $\tilde{\vb{o}}\in\mathbb{R}^{d}$. We implement $\vb{F}$ with a ResNet-18~\cite{he2016deep} backbone. Denoting the feature from the final layer before softmax of ResNet as $\vb{f}\in\mathbb{R}^{512}$, we obtain $\tilde{h}=\mathtt{MLP_2}(\vb{f})$ and $\tilde{\vb{o}}=\mathtt{FC_5}(\vb{f})$, with $\mathtt{MLP_2}$ having three layers of width $[d,d,6N]$ and adopting LeakyReLu activation for its hidden layers.

\subsubsection{Training Details}
We train a shared $\vb{F}$ across the 5 camera views of H2O and 6 fixed camera views of AssemblyHands that are leveraged in our evaluation. Noticing the different camera intrinsics among different views, we align them to a canonical camera intrinsic $K_n$, whose focal length is 240 and whose principal point is the center of the image. 
To correct 3D GT hands for the 2D projections under $K_n$, we correspondingly apply a rigid extrinsic transformation to 3D GT hands, where the transformation is obtained by solving a perspective-$n$-point problem~\cite{lepetit2009ep} with the original GT 3D joints, their 2D projections and $K_n$. The transformed 3D hand joints $\overline{h}$ are used for supervising the training of $\vb{F}$.

The training loss of $\vb{F}$ has two parts:
\begin{itemize}
    \item The hand loss $L_{hand}$ as adopted from HTT~\cite{wen2023hierarchical}, which minimizes the L1-distance between $\tilde{h}$ and the GT $\overline{h}$, regarding the 2D projection and joint depth to the camera.
    \item The object loss $L_{obj}$ to compare $\tilde{\vb{o}}$ with the object taxonomy, which is computed in a similar way as ~\cref{eq:action} of the main text. Specifically, denoting the embedding of the object taxonomy as $\mathcal{O}^F=\{\overline{o}\in\mathbb{R}^d\}$, with $\overline{o}$ as the CLIP text embeddings~\cite{cherti2023reproducible,Radford2021LearningTV,schuhmann2022laionb} for object labels, the loss is written as:
        \begin{equation}
        L_{obj}=\sum_{i=1}^{N_O} w^o_i\left(||\tilde{\vb{o}}-\overline{o}_i||_1 - \log\text{Pr}(\overline{o}_i|\tilde{\vb{o}})\right)\label{eq:obj}
        \end{equation}
    where $N_O$ is the size of the taxonomy, $w^o_i$ is 1 for the GT object and 0 otherwise, and $\text{Pr}(\cdot|\cdot)$ is computed by following~\cref{eq:action_prob} of the main text.
\end{itemize}

The overall loss for $\vb{F}$ sums them up: $L_F=L_{hand}+L_{obj}$. 

We initialize the ResNet-18 backbone with weights pre-trained on ImageNet and freeze the layers of batch normalization. The remaining weights are updated in backpropagation. We adopt the training augmentation strategy from HTT~\cite{wen2023hierarchical}, and use the AdamW optimizer~\cite{loshchilov2017decoupled}, whose learning rate is initialized as $10^{-4}$ and halved every 20 epochs. We train for 100 epochs with a batch size of 256.

\subsection{Setup of $\vb{E}^{PO}$ (\cref{sec:method_training} of Main Text)}
\subsubsection{Network} We implement $\vb{E}^{PO}$ by 2 transformer encoder layers, where the configuration of each layer is the same as that of $\vb{E}^P,\vb{E}^A$. $\vb{E}^{PO}$ takes as input a sequence of $t+1$ tokens $(\tilde{\mu}^{O},\tilde{\vb{o}}_1,...,\tilde{\vb{o}}_t)$, with $\tilde{\mu}^{O}\in\mathbb{R}^d$ as the trainable token to aggregate the clip-wise object feature. The clip-wise object feature $\vb{\omega}$ is then obtained as the first output token of $\vb{E}^{PO}$.

\subsubsection{Training Details} Based on the pre-trained $\vb{F}$, we train $\vb{E}^{PO}$ across \textit{cam0,1} of H2O and \textit{v3} of AssemblyHands, where these views are leveraged in our implementation of HTT (\cref{sec:exp_jointrp} of main text). Note that we do not update the pre-trained $\vb{F}$ in this process.

For training loss, we adopt the $L_{obj}$ defined in~\cref{eq:obj} by replacing the frame-wise $\tilde{\vb{o}}$ with the clip-wise $\vb{\omega}$. 

For training augmentation, we add random noise sampled from $N(\vec{\mathbf{0}},0.01\mathbf{I})$ to the input $\tilde{\vb{o}}$.

We train with the AdamW~\cite{loshchilov2017decoupled} optimizer, and respectively set the learning rate to $10^{-4}$, weight decay to 0.01, batch size to 256, and number of training epochs to 30.

\section{Datasets, Evaluation Setup, and Baseline Implementation}\label{sec:supp_eval_setup}

\begin{figure*}[!t]
\centering
\includegraphics[width=0.99\linewidth]{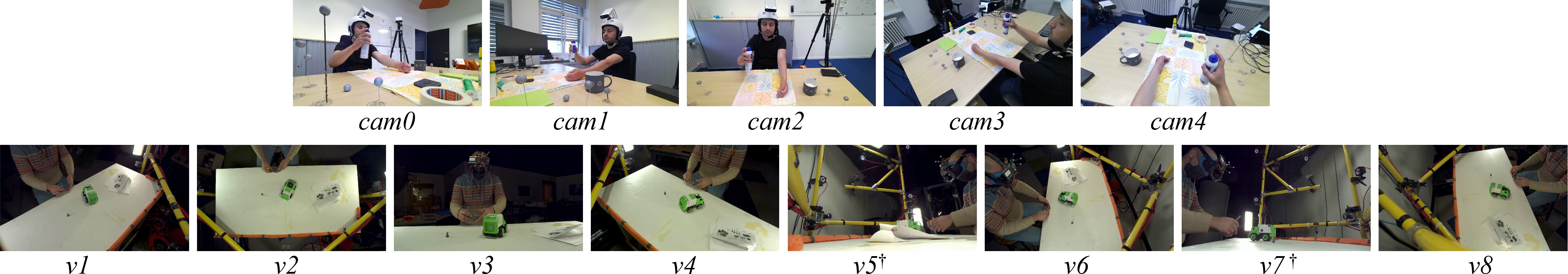}
\caption{All camera views of H2O (upper row), and fixed camera views of the Assembly101/AssemblyHands (lower row). Views denoted in \dag (\ie \textit{v5,v7}) are not leveraged due to frequent severe occlusion.}\label{fig:supp_views}
\end{figure*}

\begin{figure*}[!t]
\centering
\includegraphics[width=0.99\linewidth]{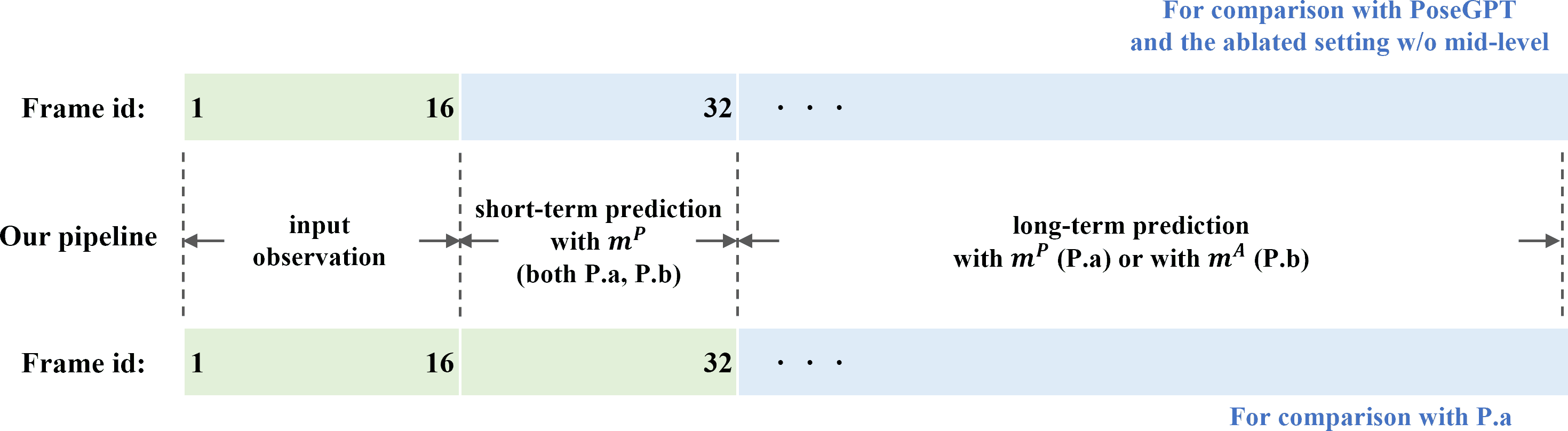}
\caption{Illustration of the evaluation setup for prediction. We focus on frames denoted in blue, to evaluate the full predictions (upper row of the figure) and to further examine the long-term prediction (lower row). For FID, the concatenation of input and prediction are fed into $\vb{E}^{FID}$, where for frames denoted in green and blue we respectively refer to the GT and network output.
}\label{fig:supp_fid}
\end{figure*}


\subsection{Camera Views of Datasets (\cref{sec:dataset} of main text)}
In~\cref{fig:supp_views}, we visualize all camera views of H2O~\cite{kwon2021h2o} and all fixed camera views of Assembly101~\cite{sener2022assembly101}/ AssemblyHands~\cite{ohkawa2023assemblyhands} datasets.
For evaluation, on H2O we refer to all camera views, including the fixed views of \textit{cam0-3}, and the egocentric view of \text{cam4} with moderate head motion; 
on AssemblyHands we refer to fixed views of \textit{v1-4,v6,v8}, but ignore \textit{v5,v7} due to the frequent severe occlusion.

\subsection{Evaluation Setup and Metrics for Prediction (\cref{sec:metrics} of main text)}
\subsubsection{FID} 
FID computation demands a feature space representation of action sequences, that generally comes from recognition models \cite{FID17}.
Without canonical models to rely on, we train action recognition network $\vb{E}^{FID}$ respectively on the training split of H2O and AssemblyHands, which aggregates a latent global feature from the raw pose sequences for action recognition. 
Note that the evaluation splits are unseen during the training of $\vb{E}^{FID}$.

The $\vb{E}^{FID}$ consists of 9 transformer encoder layers, with the configuration of each layer the same as that of $\vb{E}^{P},\vb{E}^{A}$. Specifically, the input sequence of $\vb{E}^{FID}$ has $f+1$ tokens $(\tilde{\mu}^{F}, \tilde{\vb{\gamma}}^F_1, ..., \tilde{\vb{\gamma}}^F_f)$, with $f$ as the number of frames. $\tilde{\mu}^{F}$ is the trainable token to capture the global feature, and $\tilde{\vb{\gamma}}^F_i$ encodes the input hand pose $\tilde{\vb{H}}_i$. When evaluating with FID, we leverage the first output token $\vb{\alpha}^F$ of $\vb{E}^{FID}$. 

We train $\vb{E}^{FID}$ by constraining $\vb{\alpha}^F$ with the action recognition loss defined in~\cref{eq:action} of main text:
Specifically, for H2O we compare its action taxonomy with $\vb{\alpha}^F$; for AssemblyHands we leverage the verbs only, as we notice the noun object labels provide shortcuts for distinguishing actions in the Assembly datasets~\cite{sener2022assembly101,ohkawa2023assemblyhands} without telling different hand motions.

\subsubsection{APD} Given the set of generated samples for hand pose sequence $\{h^i\in\mathbb{R}^{F\times N \times 3} | \\ i\in[1,K]\}$, with $K$ and $F$ as the number of samples and frames, we compute the Average Pairwise Diversity for the sample set using the following formula:
\begin{equation}
\begin{split}
APD =\frac{1}{K(K-1)}\sum_{\substack{a,b\in [1,K], a\neq b}} dist(h^a,h^b) \\
\text{with} \ \ dist(h^a,h^b) = \sum_{f=1}^F\sum_{n=1}^N ||h^a[f,n]-h^b[f,n]||
\end{split}
\end{equation}
where $h^i[f,n]$ denotes the 3D coordinates of $n$-\textit{th} joint in the $f$-\textit{th} frame for the  sample $h^i$.

\subsubsection{Setup (\cref{tab:prediction_posegpt,tab:prediction_ponly,tab:prediction_flatten} of main text)} 
In~\cref{fig:supp_fid}, we visualize the detailed computation of FID and APD, for the evaluation of full prediction (\ie comparison with PoseGPT~\cite{lucas2022posegpt} and the flattened setup of w/o mid-level) and the further examination of long-term prediction (\ie comparison between paths P.a and P.b). 
Specifically, the comparison with P.a replaces P.a with GT for short-term prediction (17-32 frames), to ensure a clear separation of effects from P.a and P.b.

\subsection{Implementation of HTT~\cite{wen2023hierarchical} (\cref{sec:exp_jointrp} of Main Text)}

We adopt the official code of HTT, and make the following modifications for fair comparison in terms of input/output and network size: 
\begin{itemize}
\item We train a shared network on the camera views of \textit{cam0,1} of H2O and \textit{v3} of AssemblyHands.
\item The per-frame input of the pose block is written as $\mathtt{FC_6}([\vb{f},\mathtt{FC_7}(\tilde{\vb{o}}),\mathtt{FC_8}(\tilde{h})])$, with $\vb{f},\tilde{\vb{o}},\tilde{h}$ as the image-feature, object feature and hand pose outputted by the pre-trained image-based estimator $\vb{F}$ (\cref{sec:supp_ibe}). $\mathtt{FC_6},\mathtt{FC_7},\mathtt{FC_8}$ output 512-dimensional features.
\item To align with our $\vb{E}^P,\vb{E}^A$, both pose block and action block consist of 9 transformer encoder layers. The pose block and action block respectively work on time spans of $t=16$ and $T=256$ consecutive frames at 30 fps.
\item For pose block, the object classification loss follows our~\cref{eq:obj} to compare the output with the CLIP~\cite{cherti2023reproducible,Radford2021LearningTV,schuhmann2022laionb} text embeddings of the object taxonomy. 
For the action block, the action classification loss follows~\cref{eq:action} of our main text to compare with the CLIP text embeddings of the action taxonomy. 
\item $\vb{F}$ is not updated in training the HTT.
\end{itemize}

\begin{table}[!t]
\centering
\caption{Pose estimation and action recognition results on all camera views leveraged in evaluation, for the H2O-Val, H2O-Test, and AssemblyHands-Val splits. For hand pose estimation we report MPJPE-RA/-PA in $mm$ for (left,right) hands respectively.
$*$ denotes training views leveraged for HTT.}\label{tab:supp_recognition}

\begin{subtable}[t]{0.49\textwidth}
\centering
\caption{H2O-Val}
\resizebox{0.99\linewidth}{!}{
\begin{tabular}{C{1cm}|C{2cm}|C{2cm}|C{2cm}|C{2cm}}
\hline 
\multicolumn{2}{c|}{} & Resnet-18($\vb{F}$) & HTT~\cite{wen2023hierarchical} & Ours \\
\hline
\multirow{3}{*}{cam0$^*$} & MPJPE-RA$\downarrow$ &30.5,26.6&\textbf{29.1,25.5}&29.9,26.4\\
& MPJPE-PA$\downarrow$ & 10.1,9.9 & \textbf{9.0},9.4 & 9.6,\textbf{9.3}\\ 
& Action Acc.$\uparrow$ & - & \textbf{80.33} & 63.11\\
\hline

\multirow{3}{*}{cam1$^*$} & MPJPE-RA$\downarrow$& 29.4,31.1 & \textbf{28.5},31.2 & 29.1,\textbf{30.7} \\
& MPJPE-PA$\downarrow$ & 10.2,10.9 & \textbf{9.3,10.3} & 9.7,10.4\\ 
& Action Acc.$\uparrow$ & - &\textbf{84.43}& 61.48\\
\hline

\multirow{3}{*}{cam2} & MPJPE-RA$\downarrow$  & 25.4,25.6 & 27.8,27.6 & \textbf{24.9,25.5} \\
& MPJPE-PA$\downarrow$& 9.3,8.9&9.8,8.9&\textbf{9.0,8.4}\\ 
& Action Acc.$\uparrow$ & - &\textbf{78.69} & 69.67\\
\hline

\multirow{3}{*}{cam3} & MPJPE-RA$\downarrow$ & 22.1,26.6&98.3,96.8 & \textbf{21.6,26.2} \\
& MPJPE-PA$\downarrow$ & 8.6,8.9 & 22.6,24.6 & \textbf{8.2,8.4}\\ 
& Action Acc.$\uparrow$ & - & 14.75 & \textbf{66.39}\\
\hline

\multirow{3}{*}{cam4} & MPJPE-RA$\downarrow$ & 19.1,20.7 & 120.6,139.4 & \textbf{18.8,20.4} \\
& MPJPE-PA$\downarrow$ & 7.4,7.1 & 30.1,31.3 & \textbf{7.0,6.7}\\ 
& Action Acc.$\uparrow$ & - & 6.56 & \textbf{77.87}\\
\hline
\end{tabular}}
\end{subtable}
\begin{subtable}[t]{0.49\textwidth}
\centering
\caption{H2O-Test}
\resizebox{0.99\linewidth}{!}{
\begin{tabular}{C{1cm}|C{2cm}|C{2cm}|C{2cm}|C{2cm}}
\hline 
\multicolumn{2}{c|}{} & Resnet-18($\vb{F}$) & HTT~\cite{wen2023hierarchical} & Ours \\
\hline
\multirow{3}{*}{cam0$^*$} & MPJPE-RA$\downarrow$ & 27.0,25.6 &26.9,\textbf{24.1} & \textbf{26.5},25.3 \\
& MPJPE-PA$\downarrow$ & 7.8,10.6 & \textbf{7.3},10.4&7.4,\textbf{10.3}\\ 
& Action Acc.$\uparrow$ & - & \textbf{85.12}&59.92\\
\hline

\multirow{3}{*}{cam1$^*$} & MPJPE-RA$\downarrow$ & 27.0,32.4 & \textbf{19.5,32.1} & 19.8,\textbf{32.1}\\
& MPJPE-PA$\downarrow$ & 7.3,11.6&\textbf{7.0},11.4&\textbf{7.0,11.3}\\ 
& Action Acc.$\uparrow$ & - & \textbf{88.84} & 65.70\\
\hline

\multirow{3}{*}{cam2} & MPJPE-RA$\downarrow$ &19.2,24.8&20.1,25.4&\textbf{18.9,24.5}\\
& MPJPE-PA$\downarrow$ &6.9,10.6&7.4,11.0&\textbf{6.6,10.3}\\
& Action Acc.$\uparrow$ & -&\textbf{73.55}&68.18\\
\hline

\multirow{3}{*}{cam3} & MPJPE-RA$\downarrow$ & 19.3,28.1 & 80.5,102.2&\textbf{19.0,27.8}\\
& MPJPE-PA$\downarrow$ & 7.3,10.8 & 20.8,27.2 &\textbf{7.1,10.5}\\ 
& Action Acc.$\uparrow$ & - & 12.40&\textbf{57.85}\\
\hline

\multirow{3}{*}{cam4} & MPJPE-RA$\downarrow$ &  18.4,21.4&101.2,137.8&\textbf{17.9,21.0}\\
& MPJPE-PA$\downarrow$ & 6.8,9,4&28.5,33.8&\textbf{6.4,9.1} \\
& Action Acc.$\uparrow$ &-&2.89&\textbf{57.85}\\
\hline
\end{tabular}}
\end{subtable}

\begin{subtable}[t]{0.94\textwidth}
\centering
\caption{AssemblyHands-Val}
\resizebox{0.99\linewidth}{!}{
\begin{tabular}{C{1cm}|C{2cm}|C{2cm}|C{2cm}|C{2cm}|C{1cm}|C{2cm}|C{2cm}|C{2cm}|C{2cm}}
\hline 
\multicolumn{2}{c|}{} & Resnet-18($\vb{F}$) & HTT~\cite{wen2023hierarchical} & Ours & \multicolumn{2}{c|}{} & Resnet-18($\vb{F}$) & HTT~\cite{wen2023hierarchical} & Ours\\
\hline

\multirow{3}{*}{v1} & MPJPE-RA$\downarrow$ & 35.4,22.7& 55.6,39.0&\textbf{35.1,22.4} & \multirow{3}{*}{v2} & MPJPE-RA$\downarrow$ & 31.6,28.4 & 63.8,78.2 & \textbf{31.3,28.0} \\
& MPJPE-PA$\downarrow$ & 12.0,10.8 & 17.2,14.2& \textbf{11.7,10.4}&& MPJPE-PA$\downarrow$ & 11.6,11.3 & 20.1,19.4 & \textbf{11.2,11.0}\\ 
& Action Acc.$\uparrow$ & -&16.55&\textbf{36.01}&& Action Acc.$\uparrow$ & - & 9.00 & \textbf{35.52}\\
\hline

\multirow{3}{*}{v3$^*$} & MPJPE-RA$\downarrow$ & 27.5,27.2&\textbf{26.7},27.3&27.3,\textbf{26.9} & 
\multirow{3}{*}{v4} & MPJPE-RA$\downarrow$ & 24.0,26.6 & 34.5,35.5 & \textbf{23.9,26.4}\\
& MPJPE-PA$\downarrow$ & 12.2,12.0&12.3,12.1&\textbf{11.9,11.7} & 
& MPJPE-PA$\downarrow$ & 11.0,11.6 & 14.4,14.5 & \textbf{10.8,11.3} \\ 
& Action Acc.$\uparrow$ & -&\textbf{39.42}&34.79 & 
& Action Acc.$\uparrow$ & - & 19.95 & \textbf{36.98} \\
\hline

\multirow{3}{*}{v6} & MPJPE-RA$\downarrow$ & 35.8,25.8 & 76.6,47.1 & \textbf{35.5,25.6}  & 
\multirow{3}{*}{v8} & MPJPE-RA$\downarrow$ & 26.1,30.4&91.3,88.5&\textbf{25.9,30.0}\\
& MPJPE-PA$\downarrow$ & 13.0,11.5 & 21.2,17.5 & \textbf{12.6,11.2} &
& MPJPE-PA$\downarrow$ &11.8,12.3&24.2,27.3&\textbf{11.5,11.8}\\
& Action Acc.$\uparrow$ & - & 12.90 & \textbf{33.58} &
& Action Acc.$\uparrow$ &  - &9.98&\textbf{36.74}\\
\hline
\end{tabular}}
\end{subtable}

\end{table}

\subsection{Implementation of PoseGPT~\cite{lucas2022posegpt} (\cref{sec:exp_jointrp} of Main Text)}

We adopt the official code of PoseGPT, and make the following modifications for fair comparison in terms of dataset and  encoding for pose and action: 
\begin{itemize}
\item We train the network by combining the training sets of H2O, Assembly101, and AssemblyHands into one set. 
\item We adopt our $\vb{H}$ for frame-wise hand pose representation, except that we obtain $p^L,p^R$ by aligning only the root position with the predefined template~\cite{javier2017mano}, and therefore having $v^L,v^R\in\mathbb{R}^3$ represent the relative translation. 
The reasons for this root-aligned encoding are that: (1) we observe that our original encoding makes the prediction of PoseGPT suffering from severe jittering; (2) we notice that PoseGPT represents local body-pose via root-aligned global rotations (\cf PoseGPT~\cite[Sec. 3.1]{lucas2022posegpt}), which is analogous to our root aligned hand-pose representation.
\item We represent the prescribed action as its CLIP~\cite{cherti2023reproducible,Radford2021LearningTV,schuhmann2022laionb} text embedding.
\end{itemize}
We first train the VQ-VAE $E,D$ for 1200 epochs until convergence, based on which we train the generative model $G$ and observe saturated performance after 30 epochs.

\section{Additional Results for Recognition}\label{sec:supp_recognition}
\subsection{Complete Version for \cref{tab:recognition} of Main Text}

To supplement the discussion in~\cref{sec:exp_jointrp,tab:recognition} of the main text, we report in~\cref{tab:supp_recognition} the complete results on all camera views in evaluation, for the H2O-Val, H2O-Test and AssemblyHands-Val splits.
The conclusions follow that of the main text, \ie, our results are robustly accurate across different views and datasets.

\subsection{Additional Qualitative Results for Hand Pose Refinement (\cref{fig:quali_recognition} of Main Text)}
\supp{We provide more qualitative comparison with HTT~\cite{wen2023hierarchical} and image-based estimator $\vb{F}$ in the supplementary video. Results show that our G-HTT mitigates jittering present in the input hand motion, and achieves robust generalization across different datasets and camera views.}

\section{Additional Qualitative Results of Motion Prediction}\label{sec:supp_prediction}
\supp{We provide qualitative cases in the supplementary video, to examine motion prediction from the following aspects:} 
\begin{itemize}

\item \textbf{Comparison with baseline methods}, adding to \cref{fig:quali_prediction} of main text.

\item \textbf{Prediction diversity.} Our model generates diverse motions for a specified action.

\item \textbf{Prediction from long-term observation.} Given a relatively long input sequence, our results show globally action-aware and locally consistent motion.

\item \textbf{Prediction from image-based estimation.} 
Directly processing a video input, our results again show action-aware and consistent motion.

\item \textbf{Prediction from Assembly101 annotation.} The noisy Assembly101 pose annotations can still be recognized by our model and extended into plausible future motion.

\end{itemize}

\section{Discussion (\cref{sec:exp_others} of Main Text)}\label{sec:supp_discussion}

\subsection{Training with Assembly101}
We construct the ablated setting by training G-HTT only on H2O and AssemblyHands, excluding the much larger and yet lower-quality Assembly101.
We report the comparison of action recognition and motion prediction respectively in~\cref{tab:woa101_recognition} and~\cref{tab:woa101_prediction}. 
Although the ablated setting predicts more plausible motion on the AssemblyHands, by including also the large-scale Assembly101, the network shows enhanced robustness and performance generalization: it outperforms the ablated setting by a significant margin for motion prediction on H2O, and for action recognition on both datasets.

\subsection{Predicting Action Transition}
\supp{In the supplementary video, we provide two examples of our predicted motion, which depicts the action transition from \textit{place spray} to \textit{apply spray}, and that from \textit{grab cappuccino} to \textit{take out cappuccino}.}

\begin{table}[!t]
\centering
\caption{Comparison of action recognition, between ours and the ablated setup trained without leveraging Assembly101.}\label{tab:woa101_recognition}
\begin{subtable}[t]{0.9\textwidth}
\centering
\caption{H2O-Val}
\resizebox{0.8\linewidth}{!}{
\begin{tabular}{c|ccccc}
\hline 
Action Acc.&cam0&cam1&cam2&cam3&cam4\\
\hline
\tabincell{c}{Training w/o Assembly101} & 60.66 & \textbf{61.48} & 63.11 & 57.38 & 76.23\\
Ours & \textbf{63.11} & \textbf{61.48} & \textbf{69.67} & \textbf{66.36} & \textbf{77.87}\\
\hline
\end{tabular}}
\end{subtable}

\begin{subtable}[t]{0.9\textwidth}
\centering
\caption{H2O-Test}
\resizebox{0.8\linewidth}{!}{
\begin{tabular}{c|ccccc}
\hline 
Action Acc.&cam0&cam1&cam2&cam3&cam4\\
\hline
\tabincell{c}{Training w/o Assembly101} & 55.79&58.26&61.16&55.79&\textbf{57.85}\\
Ours & \textbf{59.92} & \textbf{65.70} & \textbf{68.18} & \textbf{57.85} & \textbf{57.85}\\
\hline
\end{tabular}}
\end{subtable}

\begin{subtable}[t]{0.9\textwidth}
\centering
\caption{AssemblyHands-Val}
\resizebox{0.8\linewidth}{!}{
\begin{tabular}{c|cccccc}
\hline 
Action Acc.&v1&v2&v3&v4&v6&v8\\
\hline
\tabincell{c}{Training w/o Assembly101} & 28.47&28.22&27.25&27.74&26.76&27.01\\
Ours & \textbf{36.01} & \textbf{35.52} & \textbf{34.79} & \textbf{36.98} & \textbf{33.58}&\textbf{36.74}\\
\hline
\end{tabular}}
\end{subtable}
\end{table}

\begin{table}[!t]
\centering
\caption{Comparison of motion prediction, between ours and the ablated setup trained without leveraging Assembly101, on action sequences that are longer than 2 sec. Both setups refer to path P.b for motion prediction. APD  in \textit{mm} for (left,right) hands.}\label{tab:woa101_prediction}
\resizebox{0.9\linewidth}{!}{
\begin{tabular}{c|C{1.5cm}C{1.5cm}|C{1.5cm}C{1.5cm}|C{1.5cm}C{1.5cm}}
\hline 
\multirow{2}{*}{} &  \multicolumn{2}{c}{H2O-val} & \multicolumn{2}{|c}{H2O-Test} & \multicolumn{2}{|c}{AssemblyHands-Val} \\
& FID$\downarrow$ &  APD$\uparrow$ & FID$\downarrow$ & APD$\uparrow$ & FID$\downarrow$ &  APD$\uparrow$ \\
\hline
Training w/o Assembly101 & 10.72 & 20.4,25.4 & 16.29  & 16.3,34.2 & \textbf{2.69} & 24.1,36.3 \\
Ours &  \textbf{9.30} & \textbf{27.8,31.4} & \textbf{13.10} & \textbf{24.6,43.3}  & 5.21 &\textbf{33.6,40.0}\\
\hline
\end{tabular}}
\end{table}

\subsection{Predicting from Blended Input} 
Given two hand pose sequences $\tilde{\vb{H}}^a_{1:t}$ and $\tilde{\vb{H}}^b_{1:t}$ with $t=16$ frames, we derive their mid-level feature $\vb{m}^P_a,\vb{m}^P_b$, which are the $\mu^P$ output by $\vb{E}^{P}$. We then obtain $\vb{m}^P=\lambda\vb{m}^P_a+(1-\lambda)\vb{m}^P_b$, and observe the output motion decoded from $\vb{m}^P$. 

\supp{We show an example in the supplementary video, where the two input sequences $\tilde{\vb{H}}^a_{1:t}$ and $\tilde{\vb{H}}^b_{1:t}$ show tendencies of \textit{screwing} and \textit{shifting} right hand, respectively. By varying $\lambda$ from 1 to 0, the generated motion naturally interpolates between the tendencies of $\tilde{\vb{H}}^a_{1:t}$ and $\tilde{\vb{H}}^b_{1:t}$, demonstrating the regularity of the learned mid-level representation.}


\end{document}